\documentclass[10pt,twocolumn,letterpaper]{article}

\usepackage{cvpr}
\usepackage{times}
\usepackage{epsfig}
\usepackage{graphicx}
\usepackage{amsmath}
\usepackage{amssymb}
\usepackage{multirow}
\usepackage{color}
\usepackage{subfig}
\usepackage[export]{adjustbox}

\DeclareGraphicsExtensions{.pdf}

\usepackage[breaklinks=true,bookmarks=false]{hyperref}

\cvprfinalcopy 

\def\cvprPaperID{****} 

\ifcvprfinal\fi
\begin{document}

\title{One-to-Many Network for Visually Pleasing Compression Artifacts Reduction}

\author{Jun Guo\thanks{Corresponding Emails: Jun Guo (artanis.protoss@outlook.com), and Hongyang Chao (isschhy@mail.sysu.edu.cn)} \qquad Hongyang Chao\footnotemark[1]\\
School of Data and Computer Science, Sun Yat-sen University,\\
and SYSU-CMU Shunde International Joint Research Institute
}%

\maketitle
\thispagestyle{empty}

\begin{abstract}
We consider the compression artifacts reduction problem, where a compressed image is transformed into an artifact-free image. Recent approaches for this problem typically train a one-to-one mapping using a per-pixel $L_2$ loss between the outputs and the ground-truths. We point out that these approaches used to produce overly smooth results, and PSNR doesn't reflect their real performance. In this paper, we propose a one-to-many network, which measures output quality using a perceptual loss, a naturalness loss, and a JPEG loss. We also avoid grid-like artifacts during deconvolution using a ``shift-and-average" strategy. Extensive experimental results demonstrate the dramatic visual improvement of our approach over the state of the arts.
\end{abstract}
\vspace{-1em}

\section{Introduction}
Compression Artifacts Reduction is a classical problem in computer vision. This problem targets at estimating an artifact-free image from a lossily compressed image. In this age of information explosion, the number of images spreading on the Internet increases rapidly. Lossy compression (\eg, JPEG~\cite{jpeg}, WebP~\cite{webp}, and HEVC-MSP~\cite{hevc}) is inevitably adopted for saving bandwidth and storage space. However, lossy compression in its nature leads to information loss and undesired artifacts, which severely reduces the user experience. Thus, how to recover visually pleasing artifact-free images has attracted more and more attention.

\begin{figure}[t]
\vspace{-1em}
\centering
\subfloat[Results of ARCNN]{
\begin{minipage}{0.45\columnwidth}
\includegraphics[trim={0 2cm 0 3cm},clip,width=1\linewidth]{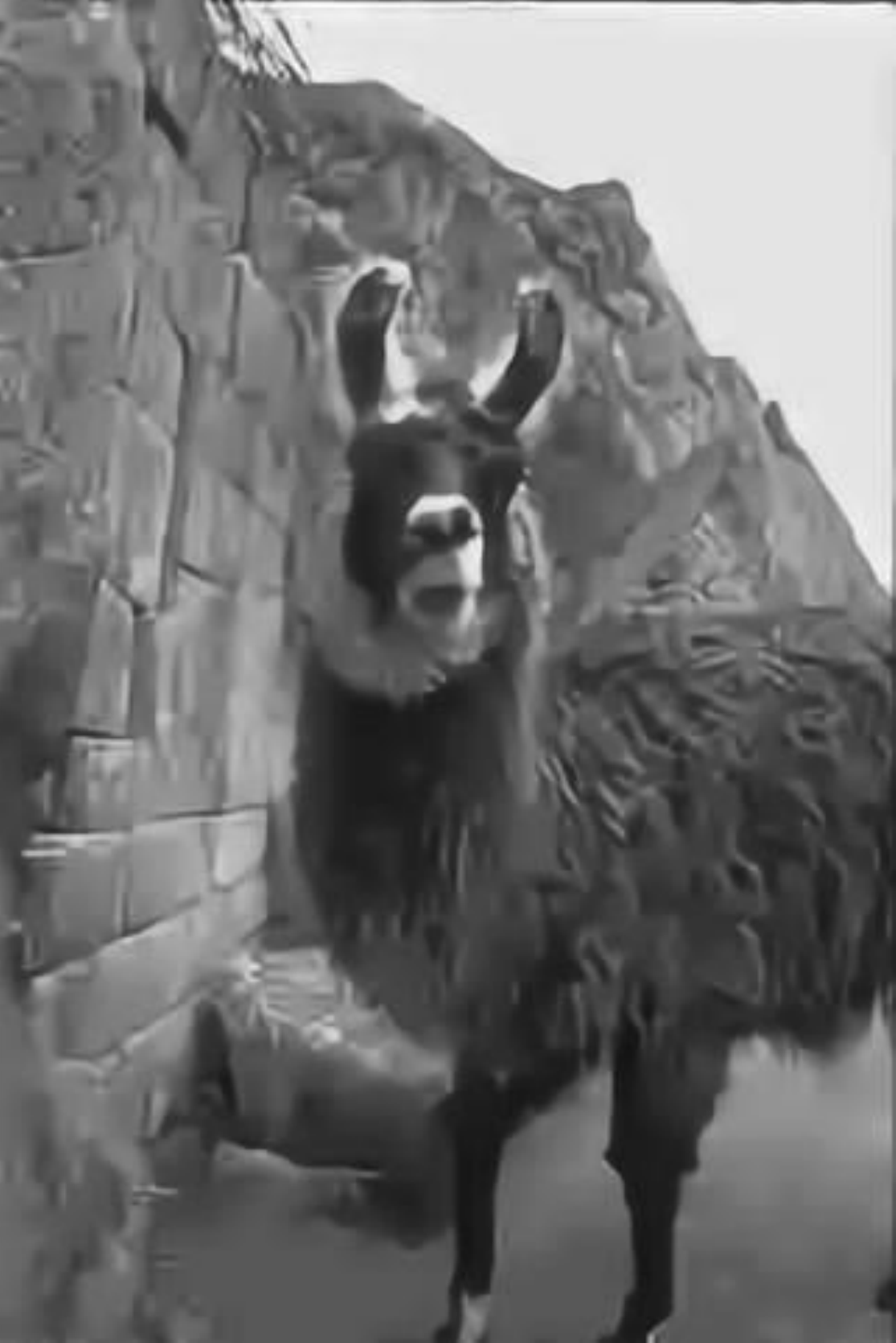}\\
\vspace{-1em}
\includegraphics[trim={1.5cm 0 1.5cm 0},clip,width=1\linewidth]{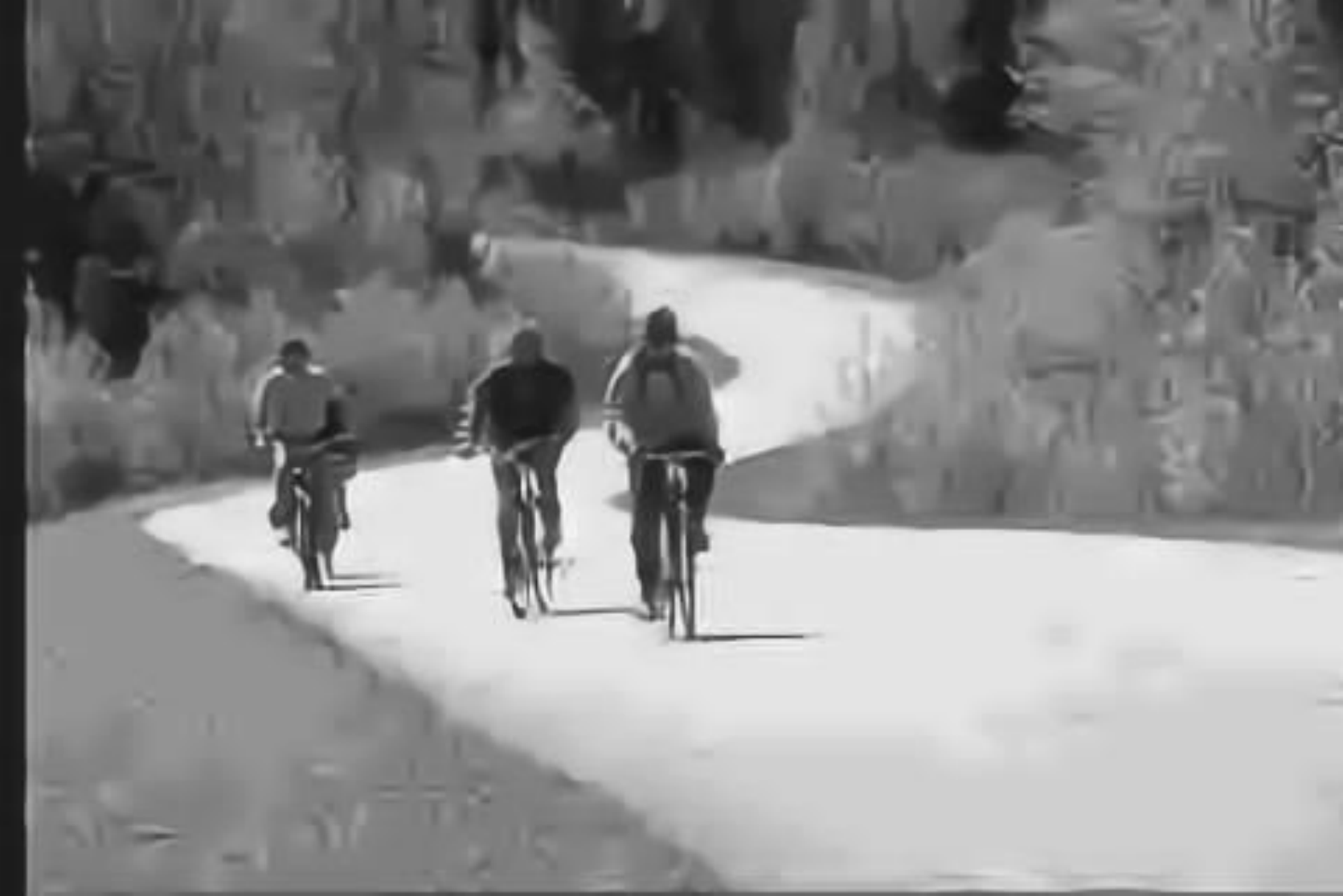}%
\label{fig:intro_arcnn}
\end{minipage}}
\hfil
\subfloat[Our results]{
\begin{minipage}{0.45\columnwidth}
\includegraphics[trim={0 2cm 0 3cm},clip,width=1\linewidth]{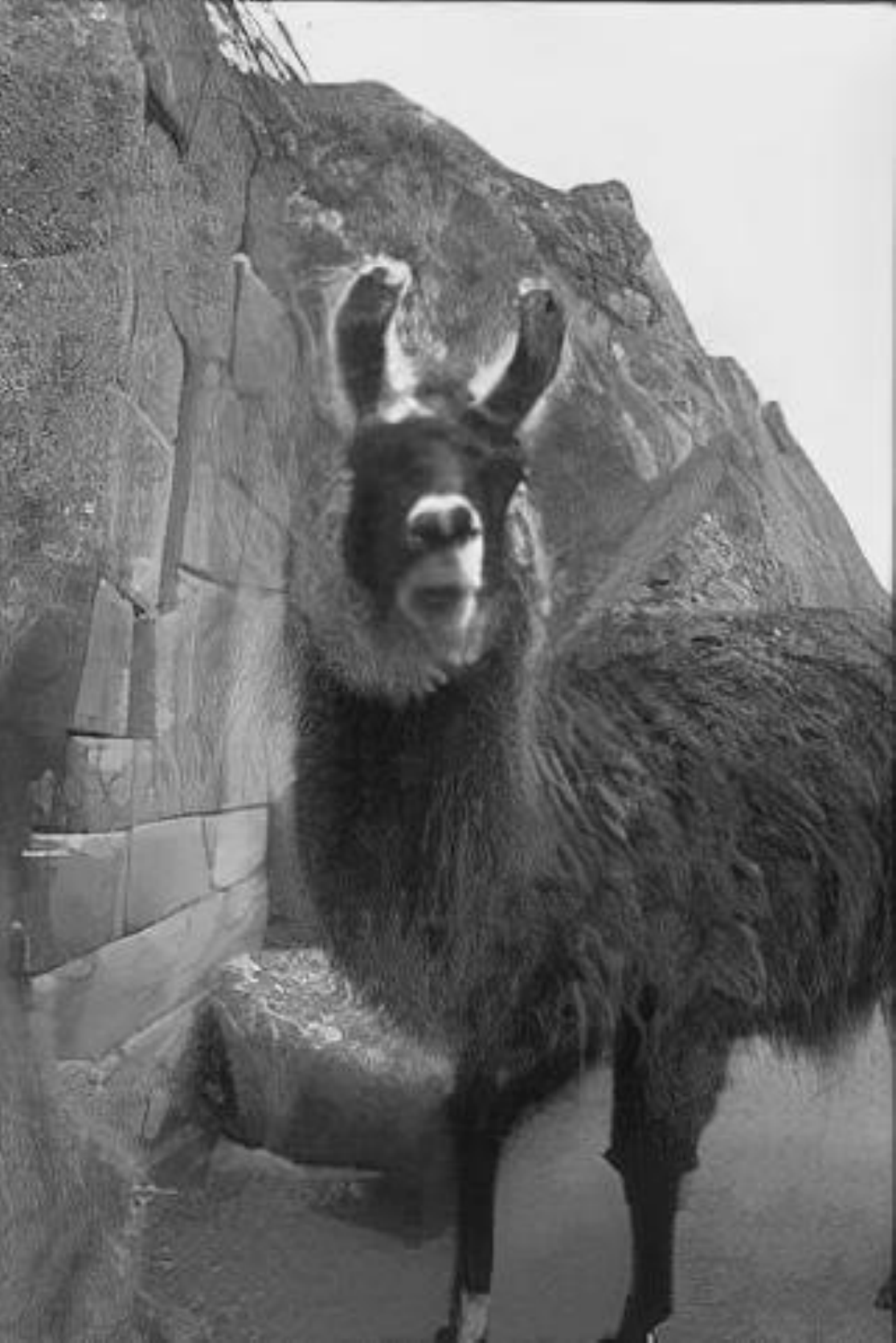}\\
\vspace{-1em}
\includegraphics[trim={1.5cm 0 1.5cm 0},clip,width=1\linewidth]{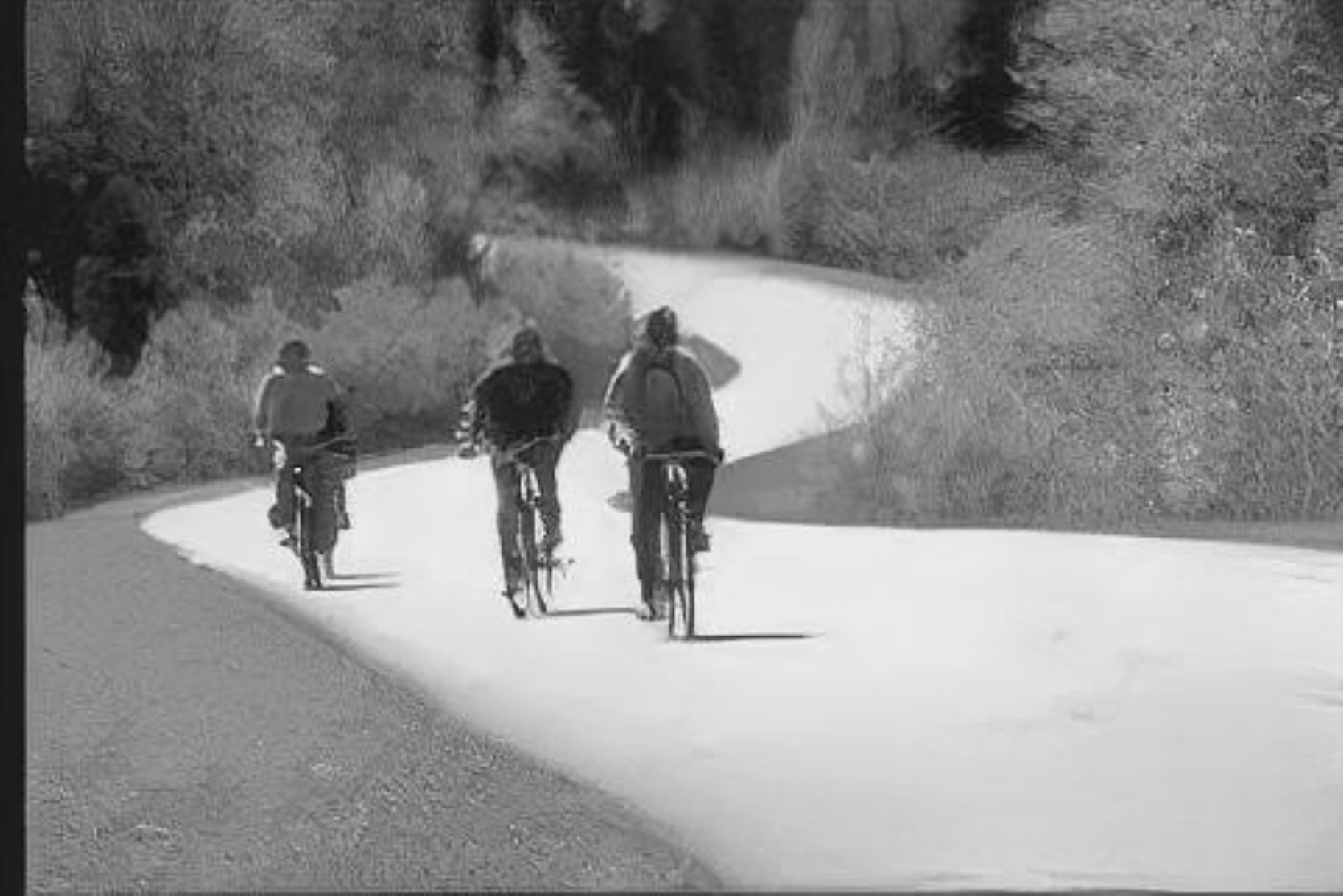}%
\label{fig:intro_ours}
\end{minipage}}
\vspace{-0.9em}
\caption{Compression artifacts reduction examples. Compared with ARCNN, our results have much richer textures; \eg, see the fur in Row 1 and the bushes in Row 2.}
\label{fig:intro}
\vspace{-1.5em}
\end{figure}

Given the fact that JPEG is the most extensively used lossy compression scheme across the world, in the following, we focus on discussing JPEG compression artifacts reduction. Various approaches have been proposed to suppress JPEG compression artifacts. Early works~\cite{review_of_postprocessing} manually developed filters to remove simple artifacts. Recently, learning-based approaches are occupying the dominant position. \cite{image_deblocking,a_learning_based,reducing_artifacts,data_driven_sparsity,inter_block_consistent,efficient_regression_priors} proposed to reconstruct artifact-free images using sparse coding. These approaches can produce sharpened images but are usually accompanied with noisy edges and unnatural regions. To date, deep learning has been proved to possess great capability for vision tasks. In particular, ARCNN~\cite{arcnn} and DDCN~\cite{ddcn} have demonstrated the power of deep convolutional neural networks (CNNs) in eliminating JPEG compression artifacts. D3~\cite{d3} casted sparse coding into a deep fully-connected network and has also obtained impressive results. Nevertheless, state-of-the-art deep learning approaches don't produce satisfactory outcomes either. The recovered images used to be overly smooth, containing significantly fewer textures when compared to the (uncompressed) ground-truth images. See Fig.~\ref{fig:intro_arcnn} for examples (Ground-truths can be found in Fig.~\ref{fig:qualitative_bsds}).

Taking an image as input, the JPEG encoder first divides it into non-overlapped coding blocks. After that, discrete cosine transform (DCT) is applied on each block, and the resultant DCT coefficients are uniformly quantified according to the JPEG quantization table. For decoding, the JPEG decoder performs inverse DCT on the quantified coefficients to obtain pixel values. It can be seen that information loss and compression artifacts are all due to quantization. Most learning-based approaches, including the aforementioned state of the arts (\eg, ARCNN, D3, and DDCN), just learn a one-to-one mapping between JPEG-compressed images and the corresponding ground-truths. Such designs have a drawback. Due to the many-to-one property of quantization, there are indeed multiple ground-truths for a compressed image. Owing to the subjective preference of images from human visual system, different people may favor different ground-truths. Therefore, it is better to develop a one-to-many mapping for recovering artifact-free images, which provide users a series of high-quality candidates, and let the users pick what they like. 

Nevertheless, measuring the output quality is a difficult task. Most existing approaches like ARCNN adopted a per-pixel $L_2$ loss, since it is straight forward and can encourage finding solutions whose pixel values are closed to ground-truths. Unfortunately, the $L_2$ loss is a convex function, so there is only one optimal solution given an input. This is contrary to the many-to-one property of quantization, and will lead to incorrect results. Consider a toy example in which $4$ different gray levels, say $1, 2, 3,$ and $4$, are all quantified to $1$. Now we are going to recover $1$$\sim$$4$ from $1$. If we learn a mapping using the $L_2$ loss, no matter how we model the mapping, at last we will find one unique solution and map $1$ to this specific value. As the mapping is trained to minimize the $L_2$ error averaged on a dataset, this solution will tend to be the ground-truth average (\eg, $\frac{1+2+3+4}{4}=2.5$), which is not any ground-truth obviously. Looking back to compression artifacts reduction, now it is also clear where those overly smooth outputs of existing approaches come from: Since each compressed image is deterministically mapped to the ground-truth average because of the per-pixel $L_2$ loss, lots of details cancel each other out during averaging, resulting in blurred regions everywhere. 

What's worse, the per-pixel loss doesn't well describe the perceptual differences between images. For instance, if we shift any of two identical images by one pixel, these two images are still perceptually similar, although they would be quite different as measured by the per-pixel loss. Recent works discovered that perceptual similarity can be better described by differences between high-level image features extracted from pretrained CNNs. This technique has been applied in feature visualization~\cite{feature_visualization}, feature inversion~\cite{feature_inversion}, style transfer~\cite{style_transfer,artistic_style}, \etc, and has succeeded in recovering semantic structures in comparison to the per-pixel loss. Nevertheless, high-level features are generally invariant to low-level details, so results from this technique usually consist of visible distortions and insufficient textures. 

On the other side, Generative Adversarial Networks (GANs)~\cite{gan} have been demonstrated to be promising in producing fine details~\cite{gan,lapgan,dcgan}. This interesting technique is usually employed to generate images that seems natural, with the naturalness measured by a binary classifier. The intuition is, if generated images are hard to be distinguished from natural images, then they should be ``natural" enough for humans. However, although significant improvements have been introduced~\cite{lapgan,dcgan}, GANs still have difficulty in generating visually pleasing semantic structures.

\begin{figure}[t]
\centering
\includegraphics[width=1\columnwidth]{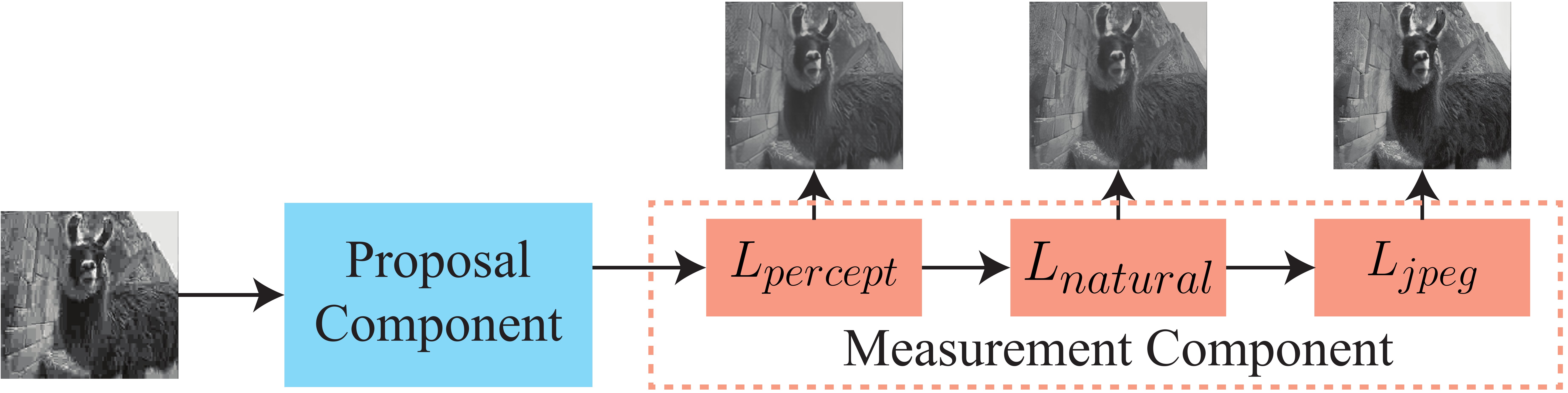}
\vspace{-1.9em}
\caption{An overview of the one-to-many network. Given a JPEG-compressed image, the proposal component generates artifact-free candidates, whose qualities are further evaluated by the measurement component.}
\label{fig:one-to-many}
\vspace{-1.5em}
\end{figure}

In this work, we combine the benefits of these two techniques within a one-to-many mapping. More specifically, we propose a one-to-many network, which is decomposed into two components -- a proposal component and a measurement component, as shown in Fig.~\ref{fig:one-to-many}. The proposal component takes a JPEG compressed image as input, and then outputs a series of artifact-free candidates. The measurement component estimates the output quality. We adopt a perceptual loss that depends on high-level features from a pretrained VGGNet~\cite{vggnet} to estimate the perceptual quality of candidates. In addition, we train a discriminative network to measure the candidate naturalness, which becomes our second loss. Meanwhile, we notice that using these two losses is still not sufficient for good results. Both of them don't respect image characteristics like color distribution. As a result, outputs from the proposal component often contain unwanted noises and have different contrast when compared with inputs or ground-truths. To resolve this issue, we further introduce a JPEG loss using the JPEG quantization table as a prior, to regularize the DCT coefficient range of outputs. Besides, we find that, when combined with highly non-convex loss functions, deconvolution usually leads to grid-like artifacts. We propose a ``shift-and-average" strategy to handle this problem. Experiments prove that our approach is able to generate results favored by humans. Compare Fig.~\ref{fig:intro_arcnn} and Fig.~\ref{fig:intro_ours} for examples.

\section{Related Works}
\subsection{Compression Artifacts Reduction}
Many approaches have been proposed to cope with compression artifacts. Early works utilized carefully chosen filters to suppress blocking and ringing artifacts. For example, Reeve and Lim~\cite{reduction_of_blocking} applied a Gaussian filter to the pixels around coding block boundaries to smooth out blocking artifacts. Besides, Chen \etal~\cite{adaptive_postfiltering} employed a low-pass filter on the DCT coefficients of adjacent coding blocks for deblocking. However, it is unlikely that such manual designs can sufficiently model compression degradation.

Recently, learning-based approaches are gradually becoming the first choice. One of the representative approaches is sparse coding~\cite{image_deblocking,a_learning_based,reducing_artifacts,data_driven_sparsity,inter_block_consistent,efficient_regression_priors}. In general, these approaches first encode an input image by a compressed-image dictionary, and then pass the sparse coefficients into an uncompressed-image dictionary for reconstruction. Sparse-coding-based approaches are usually inefficient, as they require a complicated optimization procedure. What's worse, it is quite difficult to employ end-to-end training. Thus, their performances are limited.

Neural networks date back decades~\cite{lenet}. Nowadays, deep neural networks, especially deep CNNs, have shown explosive successes in both high-level~\cite{alexnet,vggnet,learning_deconvolution} and low-level~\cite{srcnn,arcnn,d3,ddcn} vision problems. ARCNN~\cite{arcnn} demonstrated the great potential of end-to-end trained CNNs in removing various compression artifacts. DDCN~\cite{ddcn} pointed out that the $4$-layer ARCNN was insufficient to eliminate complex artifacts and thus proposed a much deeper (20-layer) architecture. D3~\cite{d3} converted sparse-coding approaches into a LISTA-based~\cite{lista} deep neural network, and obtained speed and performance gains. Both of DDCN and D3 adopted JPEG-related priors to improve reconstruction quality.

\subsection{Perceptual Loss}
A number of recent works used a perceptual loss, which is defined on high-level features extracted from pretrained CNNs, as the optimization target. Mahendran and Vedaldi~\cite{feature_inversion} inverted features from convolutional networks by minimizing the feature reconstruction loss, in order to analyze the visual information captured by different network layers. Similar optimization targets have been used in feature visualization~\cite{feature_visualization}, artistic style transfer~\cite{style_transfer,artistic_style}, and so on. In particular, Johnson \etal~\cite{perceptual_losses} trained a feed-forward network to solve the optimization problem, largely reducing the computational cost. This work is particular relevant to ours, as they have showed impressive results in image super-resolution by replacing the per-pixel $L_2$ loss with the perceptual loss during the training of a CNN. However, as discussed in the Introduction, only minimizing the perceptual loss usually leads to unsatisfactory details.

\subsection{Generative Adversarial Networks}
Starting from Goodfellow \etal~\cite{gan} who introduced GANs for generating digits, GANs have attracted significant attention in the area of image generation. In general, a GAN contains a generative network and a discriminative network. The discriminative network is trained to determine whether an image is from reality or the generative network. And the generative network is trained to improve its outputs so that they are good enough and cannot be easily distinguished from reality. Training GANs is tricky and unstable. Denton \etal~\cite{lapgan} built a Laplacian pyramid of GANs (LAPGAN) to generate natural images in a coarse to fine scheme. 
In addition, DCGAN~\cite{dcgan} proposed some good practices for training GANs. More applications of GANs can be found in \cite{deep_multi_scale,generating_images_with_recurrent,generative_image_modeling}. Concurrent with our work, Ledig \etal~\cite{srgan} also combine a VGGNet-based perceptual loss and GANs for image restoration, and have achieved impressive results.

\section{One-to-Many Network}
\subsection{Formulation}
Consider a JPEG-compressed image $Y$. Our goal is to recover from $Y$ a series of artifact-free images $F(Y)$ which are as similar as possible to the uncompressed ground-truth $X$. Note that here we only consider one ground-truth for each input. Although a compressed image may come from numerous uncompressed images, in practice, due to the available data, at most time we can only access one ground-truth for a compressed image. Nevertheless, our discussion can be easily extended to multiple ground-truths.

\subsection{Proposal Component}
Our one-to-many network contains two major components. In this sub-section we describe the proposal component, which provides a model for $F$. More specifically, we develop the mapping $F$ as a deep CNN. To enable the one-to-many property, within the network we introduce an auxiliary variable $Z$ as a hidden additional input. The network takes a compressed image $Y$ as input; at the same time it samples a $Z$ from a zero-centered normal distribution with standard deviation $1$. Both of $Y$ and $Z$ are then feed into the network for non-linear mapping. As JPEG compression is not optimal, redundant information neglected by the JPEG encoder may still be found in a compressed image. A deep CNN on $Y$ can effectively discover and utilize such information to recover details eliminated by quantization. The sampled $Z$ adds randomness to the network, encouraging it to explore and generate distinct artifact-free candidates.

\subsubsection{Network Structure}
\begin{figure}[t]
\centering
\includegraphics[width=0.85\columnwidth]{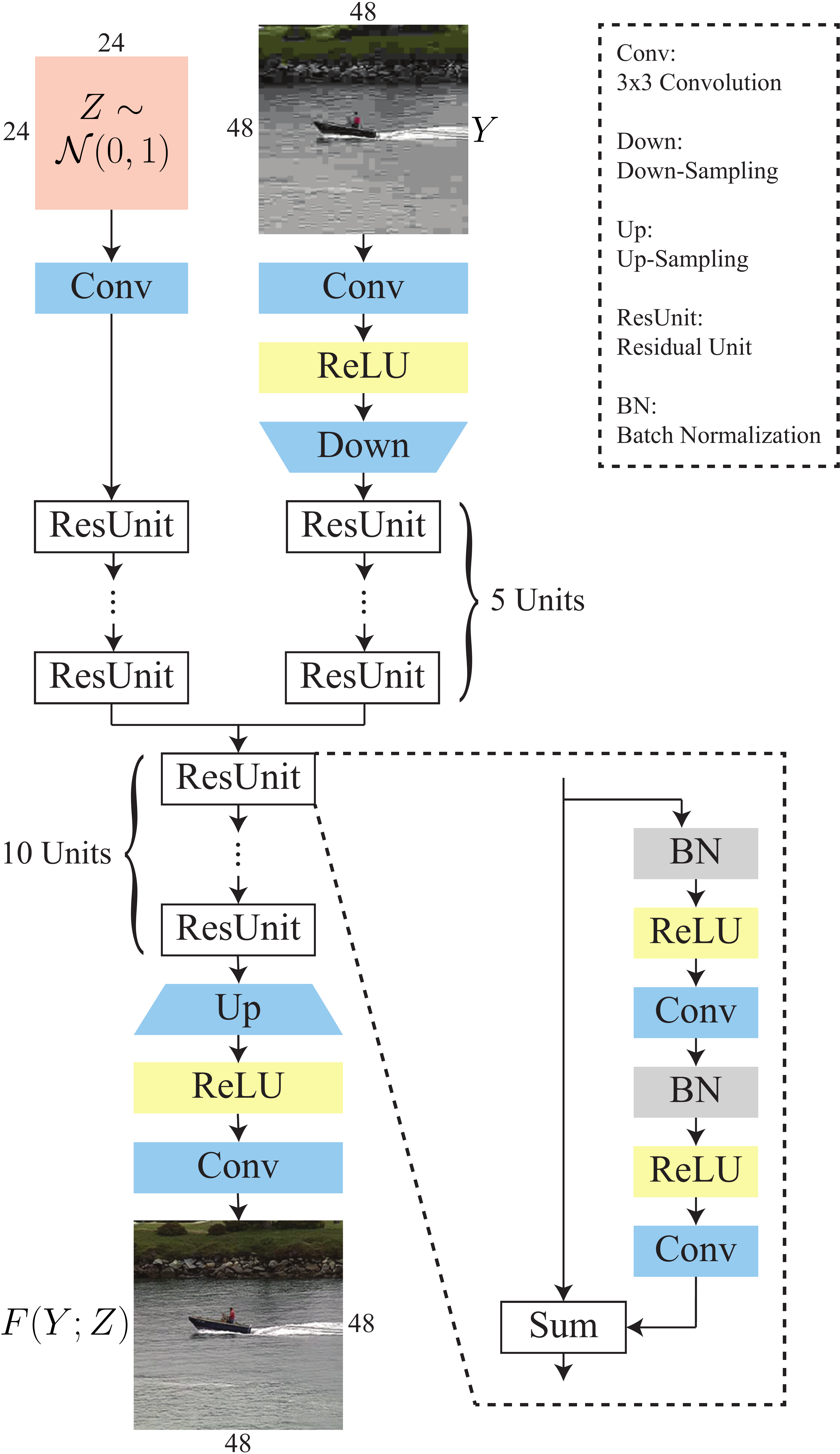}
\vspace{-1em}
\caption{The architecture of the proposal component. The filter number of the last layer is equal to the channel number of inputs. The other convolutional layers contain $64$ filters. }
\vspace{-1em}
\label{fig:proposal}
\end{figure}

The proposal component roughly follows the network structure set forth by \cite{generative_image_modeling} and \cite{perceptual_losses}. At first the compressed image $Y$ and the sampled $Z$ are given as inputs of two different branches. After that the outputs of these two branches are concatenated. On top of the concatenated feature maps, an aggregation sub-network is further performed to generate an artifact-free prediction. Fig.~\ref{fig:proposal} illustrates this component.

Very recently, skip connection, especially the identity shortcut, has become very popular in building deep neural networks. He \etal's deep residual network (ResNet)~\cite{resnet} which consists of many stacked residual units has shown state-of-the-art accuracy for several challenging recognition tasks. Our work follows their wisdom. In the proposal component, each branch contains $5$ residual units, and the aggregation sub-network comprises $10$ residual units. For the residual unit we adopt the variant proposed in He \etal's later work~\cite{identity_mappings}. More specifically, each residual unit includes two Batch Normalization~\cite{batch_normalization} layers, two ReLU~\cite{alexnet} layers, and two convolutional layers.

Before a compressed image is forwarded to the network, it is down-sampled by a stride-$2$ $4\times4$ convolutional layer. And finally the network output is up-sampled by a stride-$2$ $4\times4$ deconvolutional layer to maintain the image size. There are two benefits to networks that down-sample and then up-sample. First, due to the reduced input resolution, the computational cost is much lower (only $\frac{1}{4}$ compared with the no down-sampling version). Second, with a same filter size and a same layer number, $2$x down-sampling can increase the effective receptive field size by $2$, which is advantageous for integrating large-area spatial information.

\subsubsection{Up-Sampling}
There is no free lunch. Although down-sampling has several benefits, up-sampling is not as trivial as it first looks. Let us consider a $1$-D example, using a stride-$2$ deconvolutional layer with filter size $4$ for up-sampling. Denote the filter as $\left[w_1, w_2, w_3, w_4\right]$. Now assume we apply deconvolution on a constant input $\left[\cdots, c, c, \cdots\right]$, where $c$ is a scalar. The expected output should be constant as well. However, the actual output will be $c*\left[\cdots, w_2+w_4, w_1+w_3, w_2+w_4, w_1+w_3, \cdots\right]$. If we require the actual output to meet the expected output, then the trained filter should satisfy $w_1+w_3 = w_2+w_4$. This constraint may be implicitly learned if we use a per-pixel $L_2$ loss on top of the deconvolutional layer, because the $L_2$ loss is simple enough so that learning the filter weights is nearly a convex optimization problem. But if a highly non-convex loss is adopted, we find that a network would struggle in learning this requirement, resulting in apparent grid-like artifacts. Indeed, this kind of artifacts can be seen in results of many previous works which combined deconvolution with a complicated loss function (\eg, see Fig.~8 in~\cite{perceptual_losses}).

Note that using a different filter size won't resolve this issue (in fact, it may be worse if the filter size is odd). Instead, in this work we propose a simple strategy namely ``shift-and-average". Continuing the previous example, after we obtain the deconvolution output (denoted as $deconv$), the following two steps are performed:
\begin{enumerate}
\vspace{-0.5em}
\setlength{\itemsep}{-0.5\itemsep}
\item Duplicate $deconv$ and shift it right by $1$ pixel.
\item Average $deconv$ and the shifted version.
\vspace{-0.5em}
\end{enumerate}
\begin{figure}[t]
\centering
\includegraphics[width=0.95\columnwidth]{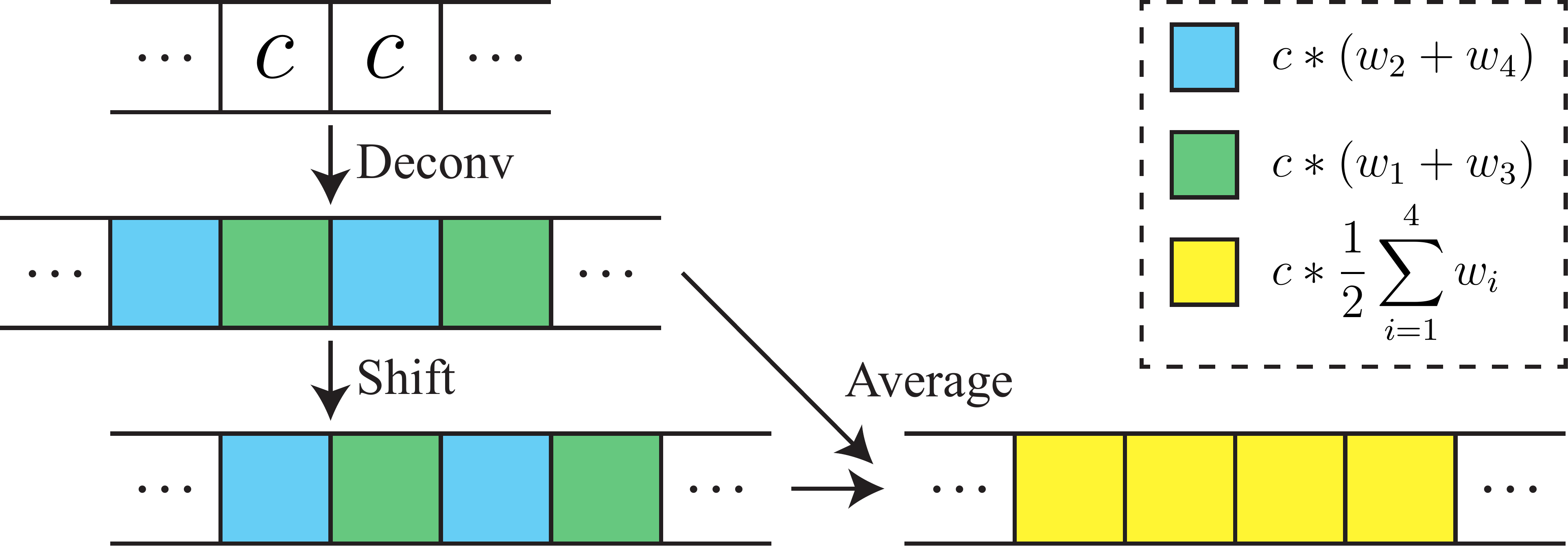}
\vspace{-0.5em}
\caption{An illustration for the shift-and-average strategy.}
\vspace{-1em}
\label{fig:upsample}
\end{figure}
Fig.~\ref{fig:upsample} provides an illustration. We can see that a constant input will result in a constant output, which is expected. This strategy can be easily extended to $2$-D data. For $2$-D stride-$2$ deconvolution, $3$ shifts (\ie, shift right, shift down, shift right and down) are performed in Step 1. In general, for stride-$N$ $2$-D deconvolution, $N^2-1$ shifts are required. Nevertheless, both steps in the proposed strategy can be efficiently parallelized, and thus run extremely fast on GPUs.

\subsection{Measurement Component}
After we obtain an output $\hat{X} = F(Y; Z)$ from the proposal component, the measurement component is adopted to estimate whether $\hat{X}$ is favored by humans. We define three loss functions for measurement.

\subsubsection{Perceptual Loss}
The perceptual loss estimates semantic similarity between $\hat{X}$ and $X$. Previous works~\cite{artistic_style,style_transfer,perceptual_losses} found that features from deep networks pretrained for image classification can well describe perceptual information. Especially, Mahendran and Vedaldi~\cite{feature_inversion} showed that, features extracted from lower layers tend to retain photographically accurate information, whereas higher-layer features are more invariant to color, texture, and shape differences. Thus, rather than advocating per-pixel matching, the perceptual loss is defined to encourage $\hat{X}$ and $X$ to share similar high-layer features:
\begin{equation}
L_{percept}(\hat{X}, X) = \frac{1}{H_\phi}\lVert\phi(\hat{X}) - \phi(X)\rVert_2^2,
\end{equation}
where $\phi$ is features computed from a network, and $H_\phi$ is the feature size. In this work, we employ the activations of the last convolutional layer of the 16-layer VGGNet~\cite{vggnet}, \ie, Layer ``relu5\_4", as $\phi$.

\subsubsection{Naturalness Loss}
The perceptual loss has high degree of geometric and photometric invariance. This is good for reconstructing semantic structures, but it also has some drawbacks. Consider a natural image and a moderately smoothed version. Minimizing the perceptual loss won't strongly favor the un-smoothed version, as normally those smoothed-out details don't have a great influence on semantic discrimination. However, we wish to recover artifact-free images as ``natural" as possible.

We introduce another loss to resolve this issue, following the spirit of GANs. We build an additional network $D$ to distinguish whether an image is generated from the proposal component $F$ or it is a natural image. Network $D$ performs binary classification and outputs the probability of an input being ``natural". We add (the negative log of) this probability on $\hat{X}$ as the second loss for the measurement component, encouraging $\hat{X}$ to have a high probability:
\begin{equation}
L_{natural}(\hat{X}) = -log(D(\hat{X})).
\label{eq:g_loss}
\end{equation}

Network $D$ needs to be trained as well. We adopt the binary entropy loss as its optimization target:
\begin{equation}
L_{D}(X, \hat{X}) = -\left(log(D(X)) + log(1 - D(\hat{X}))\right).
\label{eq:d_loss}
\end{equation}

As can be seen from Eq.~\eqref{eq:g_loss} and Eq.~\eqref{eq:d_loss}, network $F$ and network $D$ are competing against each other: network $F$ tries to generate an artifact-free image $\hat{X}$ which is hard for network $D$ to differentiate from natural images, while $D$ is trained to avoid getting fooled by $F$.

\begin{table}
\begin{center}
\caption{Comparison of DCGAN and our network $D$. ``conv" is short for convolution. Each convolutional layer except the last is followed by a Batch Normalization and a Leaky ReLU, sequentially. The filter size is always $4 \times 4$. The filter numbers are shown right after ``conv".}
\vspace{-1em}
\resizebox{0.9\columnwidth}{!}{%
\begin{tabular}{c|c|c}
\hline
& DCGAN & Network $D$\\
\hline
\multirow{2}{*}{Conv Unit 1}&                     & stride-1 conv (64)\\
                            & stride-2 conv (64)  & stride-2 conv (64)\\
\hline
\multirow{2}{*}{Conv Unit 2}&                     & stride-1 conv (128)\\
                            & stride-2 conv (128) & stride-2 conv (128)\\
\hline
\multirow{2}{*}{Conv Unit 3}&                     & stride-1 conv (256)\\
                            & stride-2 conv (256) & stride-2 conv (256)\\
\hline
\multirow{2}{*}{Conv Unit 4}&                     & stride-1 conv (512)\\
                            & stride-2 conv (512) & stride-2 conv (512)\\
\hline
Classifier                  & \multicolumn{2}{c}{Logistic Regression}\\
\hline
\end{tabular}
\label{tb:d}
}
\end{center}
\vspace{-1.5em}
\end{table}

For the structure of network $D$, we generally follow the architectural guidelines proposed by DCGAN~\cite{dcgan}, with the network depth doubled. More specifically, we also adopt $4$ convolutional units, but each unit is composed of $2$ convolutional layers instead of $1$. Every convolutional layer except the last is followed by a Batch Normalization and a Leaky ReLU~\cite{lrelu}. The outputs of the last convolutional unit are feed into a Logistic regression classifier. We despite the differences of our network $D$ and DCGAN in Table~\ref{tb:d}.\footnote{In the first version of this paper, Table~\ref{tb:d} contained a 512-node fully-connected layer between ``Logistic Regression" and the last convolutional layer. Our previous intension was to indicate that the classifier took 512-channel feature maps as its input, and at that time ``Logistic Regression" in the table just meant a sigmoid transform (not the classifier). Unfortunately, this is quite misleading. Hence, in this version we revise this part.}

\subsubsection{JPEG Loss}
Intuitively, if we adjust the contrast of an image, little semantic information will change. That is, the perceptual loss is insensitive to the color distribution of a reconstructed image $\hat{X}$. Besides, Eq.~\eqref{eq:g_loss} shows that the naturalness loss doesn't concern whether the color distributions of $\hat{X}$ and the input $Y$ match, either. Oppositely, for the purpose of compression artifacts reduction, we hope the color distribution of an input can be roughly retained. Hence, we introduce an additional JPEG-related loss to enforce this constraint.

Estimating the real color distribution is rather difficult. Fortunately, the JPEG standard is composed of various pre-defined parameters. By wisely leveraging these parameters, at least we can obtain the lower bound and the upper bound of pixel values. As aforementioned, for compression, the JPEG encoder divides the DCT coefficients of an input image by a quantization table, and then rounds the results to the nearest integer. The JPEG decoder performs decompression by multiplying back the quantization table. Thus, the relations between a compressed image $Y$ and the corresponding uncompressed image $X$ can be formulated as:
\begin{equation}
Y^{dct}(i, j) = \text{ROUND}\left(X^{dct}(i, j) / Q(i, j)\right) * Q(i, j),
\label{eq:quantization}
\end{equation}
where $X^{dct}$ and $Y^{dct}$ are the DCT coefficients of $X$ and $Y$, respectively. $Q$ is the quantization table. $i$ and $j$ are indices in the DCT domain. Eq.~\eqref{eq:quantization} implies the following DCT coefficient range constraint:
\begin{equation}
Y^{dct} - Q / 2 \le X^{dct} \le Y^{dct} + Q / 2.
\label{eq:constraint}
\end{equation}

So each recovered artifact-free image $\hat{X}$ should satisfy Eq.~\eqref{eq:constraint} as well. We propose the following JPEG loss:
\begin{equation}
L_{jpeg}(\hat{X}, Y) = \frac{1}{H_{\hat{X}}}\text{MAX}\left(\left(\hat{X}^{dct} - Y^{dct}\right)^2 - \left(\frac{Q}{2}\right)^2, 0\right)
\end{equation}
where $H_{\hat{X}}$ is the size of $\hat{X}$. As can be seen, the JPEG loss is a truncated $L_2$ loss. A reconstructed image $\hat{X}$ whose DCT coefficients fall outside the lower / upper bound (\ie, $\lvert\hat{X}^{dct} - Y^{dct}\rvert > \frac{Q}{2}$) will be penalized.

\subsection{Joint Training for the One-to-Many Network}
\label{sec:joint_training}
We merge all the aforementioned loss functions to build the measurement component:
\begin{equation}
\begin{split}
L(\hat{X}, X, Y) = &L_{percept}(\hat{X}, X) + \lambda_1 L_{natural}(\hat{X}) +\\
&\lambda_2 L_{jpeg}(\hat{X}, Y).\\
\end{split}
\label{eq:final_loss}
\end{equation}

In this paper, we always set $\lambda_1$ to $0.1$. $\lambda_2$ needs some special treatments. Note that the JPEG encoder performs quantization on each $8 \times 8$ non-overlapped coding block individually. For a patch misaligned with the coding block boundaries, we cannot obtain its DCT coefficients. Hence, we set different $\lambda_2$ values according to the given patches. Generally speaking, the training procedure for our one-to-many network is similar to the original GAN~\cite{gan}, which includes two main steps in each iteration:
\begin{enumerate}
\item Fix the proposal component $F$, optimize the discriminative network $D$ with Eq.~\eqref{eq:d_loss}.
\item Fix network $D$, optimize the proposal component $F$ with the measurement component (\ie, Eq.~\eqref{eq:final_loss}). If an input patch is aligned with the JPEG coding block boundaries, set $\lambda_2$ to $0.1$; otherwise set $\lambda_2$ to $0$.
\end{enumerate}
In the first epoch of training we only perform the second step without network $D$, \ie, network $D$ is not trained nor used in the first epoch. The reason is, at the beginning the generated images are not good, so even a trivial network $D$ can distinguish them from natural images. Feeding them to network $D$ is just a waste of computational resources.

\section{Experiments}
In this section, we conduct experiments to demonstrate the effectiveness of the proposed one-to-many network.

\paragraph{Dataset.}
In all experiments, we employ the ImageNet dataset~\cite{imagenet} for training. The validation set of the BSDS500 dataset~\cite{bsds500} is used for validation. Following the standard protocol of previous approaches, the MATLAB JPEG encoder is applied to generate JPEG-compressed images. Nevertheless, other JPEG encoders are generally acceptable as we didn't notice visible differences in resultant images. 

\paragraph{Parameter Settings.}
We roughly follow the parameter settings in DCGAN~\cite{dcgan}. We train our one-to-many network for $3$ epochs, using Adam~\cite{adam} with learning rate $1e^{-4}$ and momentum term $\beta_1 = 0.5$. The batch size is set to $16$.\footnote{In the first version of the paper, we wrote down the learning rate as $2e^{-4}$ and the batch size as $8$. These were somewhat ambiguous. In fact, we trained the network on two GPUs, with each GPU handling a batch of $8$ images. We also rescaled gradients in proportion to the batch size, for ease of hyper-parameter tuning~\cite{parallel_cnn}. Taken them into consideration, the effective learning rate and batch size should be $1e^{-4}$ and $16$ respectively when the network is trained on a single GPU without gradient rescaling.} For the Leaky ReLU, the slope of the leak is set to $0.2$. Training images are prepared as same-size patches for network inputs. All weights are initialized using He~\etal's uniform initializer~\cite{delving_deep}. During testing, the proposal component runs as a fully convolutional network~\cite{fcn} to generate full-image predictions.

\subsection{Baseline Evaluations}
The traditional metrics used to evaluate compression artifacts reduction are PSNR, SSIM~\cite{ssim}, and PSNR-B~\cite{psnrb}. All of them rely on low-level differences between pixels. Especially, PSNR is equivalent to the per-pixel $L_2$ loss. So when measured by PSNR, a model trained to minimize the per-pixel $L_2$ error should always outperform a model that minimizes Eq.~\eqref{eq:final_loss}. For fair comparison, in this experiment we replace the measurement component by the per-pixel $L_2$ loss. We also drop the auxiliary variable in the proposal component as there is only $1$ optimal solution under the $L_2$ loss. We name this variant as ``baseline" in the following.

We compare our baseline with two latest compression artifacts reduction approaches, \ie, ARCNN, and DDCN, on the test set of the BSDS500 dataset~\cite{bsds500}. We also include the latest generic image restoration framework TNRD~\cite{tnrd} for comparison. D3 is not examined here as so far there is no open code or model for evaluation. Three JPEG qualities are evaluated: $5$, $10$, and $20$. All experiments in this section are conducted on the luminance channel (in YCbCr color space), according to the protocol of previous approaches.

Table~\ref{tb:ex1} shows the quantitative results. On the whole, our baseline largely outperforms ARCNN, and TNRD on all JPEG qualities and evaluation metrics, and is on par with DDCN. In particular, our baseline performs best for Quality 5, indicating it is especially suitable for low-quality inputs, which have higher demand for good reconstruction. We emphasize that our goal of this paper is not to achieve the best PSNR / SSIM / PSNR-B results, but instead to improve the human favorability of recovered artifact-free images. Thus, we didn't add techniques like the DCT-domain priors used in D3 and DDCN to further improve PSNR.

\begin{table}
\begin{center}
\caption{Comparisons with the State of the Arts on the BSDS500 Dataset. {\color{red}Red color} indicates the best performance; {\color{blue}Blue color} indicates the second best performances.}
\vspace{-1em}
\resizebox{1\columnwidth}{!}{%
\begin{tabular}{c|c|c|c|c}
\hline
Quality & Approach & PSNR (dB) & SSIM & PSNR-B (dB)\\
\hline
\multirow{5}{*}{5}  & JPEG & 25.36 & 0.6764 & 22.91\\
                    & ARCNN & 26.72 & 0.7256 & 26.48\\
                    & TNRD & 26.81 & 0.7279 & 26.65\\
                    & DDCN & \color{blue}26.98 & \color{blue}0.7333 & \color{blue}26.76\\
                    & Baseline & \color{red}27.12 & \color{red}0.7406 & \color{red}26.87\\

\hline
\multirow{5}{*}{10} & JPEG & 27.80 & 0.7875 & 25.10\\
                    & ARCNN & 29.10 & 0.8198 & 28.73\\
                    & TNRD & 29.16 & 0.8225 & 28.81\\
                    & DDCN & \color{red}29.59 & \color{red}0.8381 & \color{red}29.18\\
                    & Baseline & \color{blue}29.56 & \color{blue}0.8352 & \color{blue}29.10\\

\hline
\multirow{5}{*}{20} & JPEG & 30.05 & 0.8671 & 27.22\\
                    & ARCNN & 31.28 & 0.8854 & 30.55\\
                    & TNRD & 31.41 & 0.8889 & 30.83\\
                    & DDCN & \color{blue}31.88 & \color{red}0.8996 & \color{red}31.10\\
                    & Baseline & \color{red}31.89 & \color{blue}0.8977 & \color{blue}31.04\\
\hline
\end{tabular}%
}
\label{tb:ex1}
\end{center}
\vspace{-2.5em}
\end{table}

\begin{figure*}[t]
\centering
\captionsetup[subfloat]{labelformat=empty}
\subfloat[Ground-truth / PSNR]{\includegraphics[trim={0 2cm 0 3cm},clip,width=0.19\linewidth]{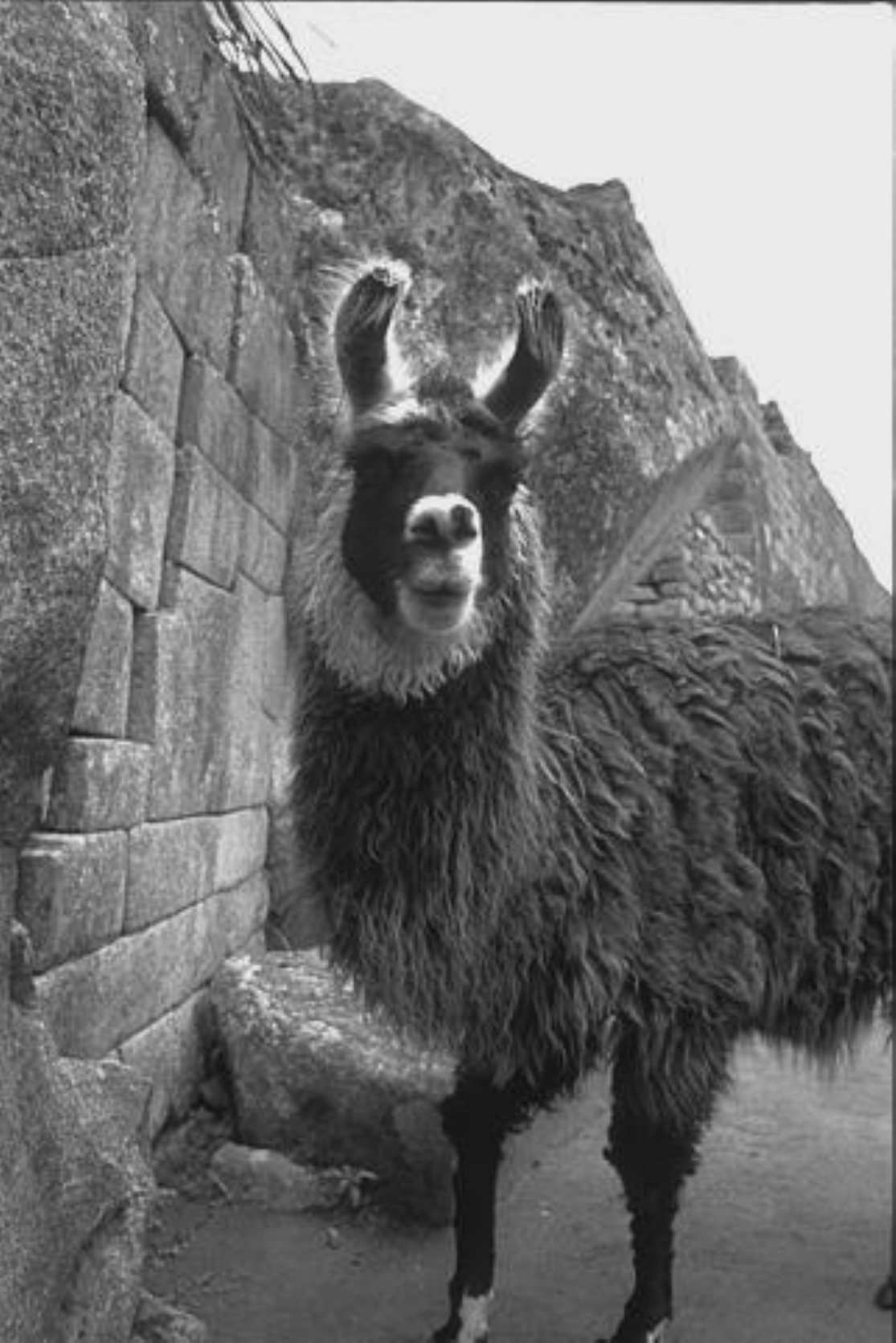}}
\hfil
\subfloat[JPEG / $24.09$]{\includegraphics[trim={0 2cm 0 3cm},clip,width=0.19\linewidth]{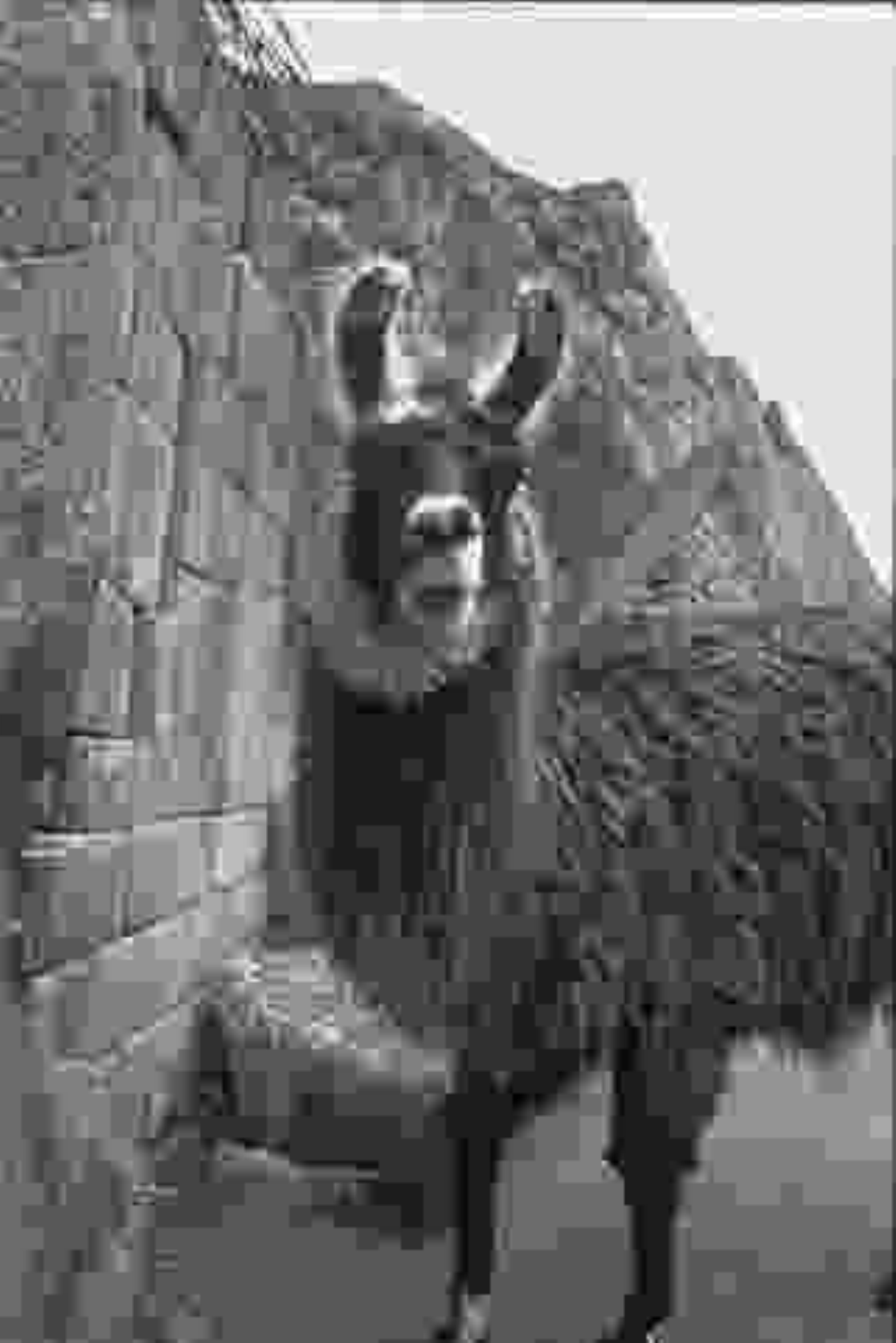}}
\hfil
\subfloat[DDCN / $25.20$]{\includegraphics[trim={0 2cm 0 3cm},clip,width=0.19\linewidth]{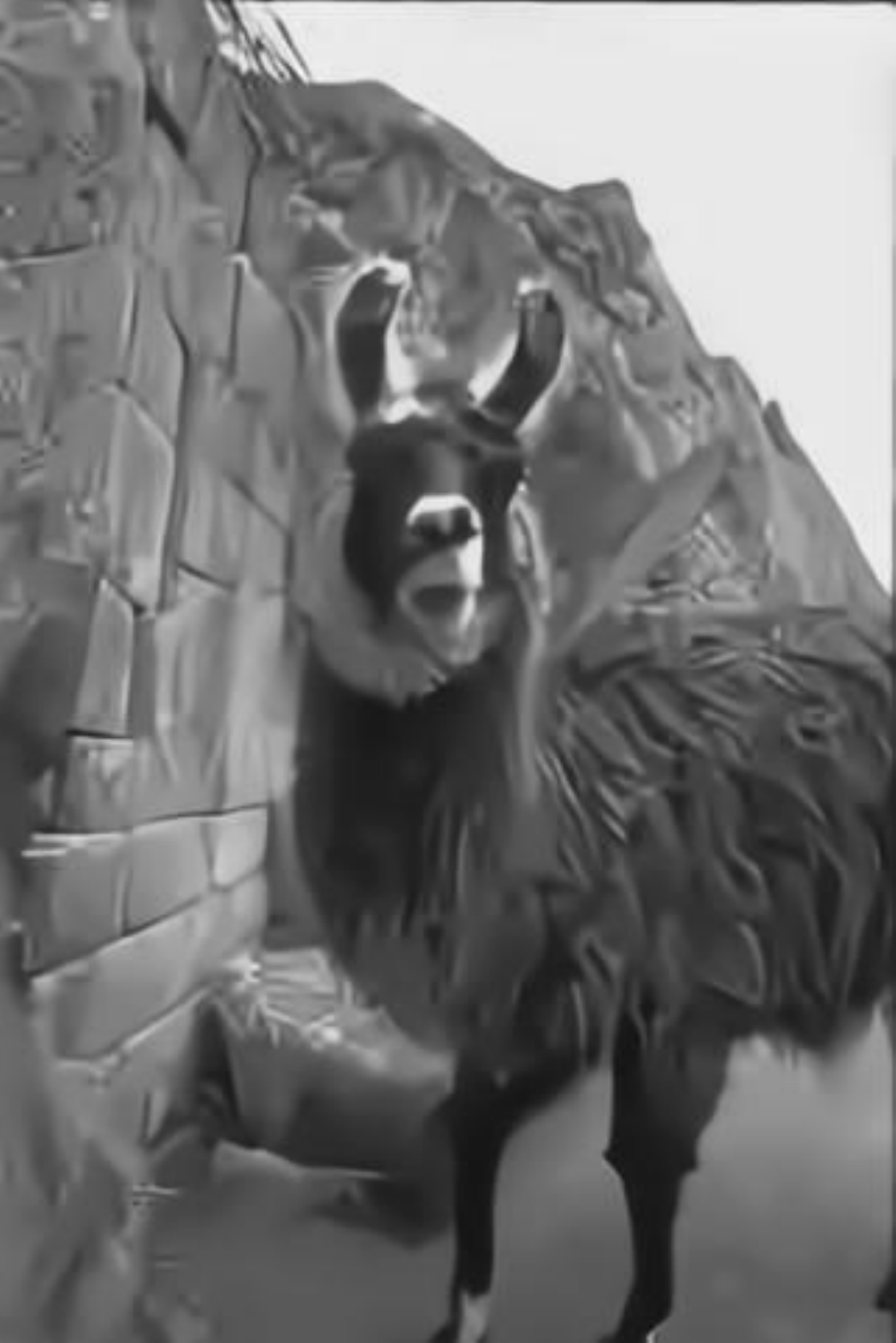}}
\hfil
\subfloat[Baseline / $25.29$]{\includegraphics[trim={0 2cm 0 3cm},clip,width=0.19\linewidth]{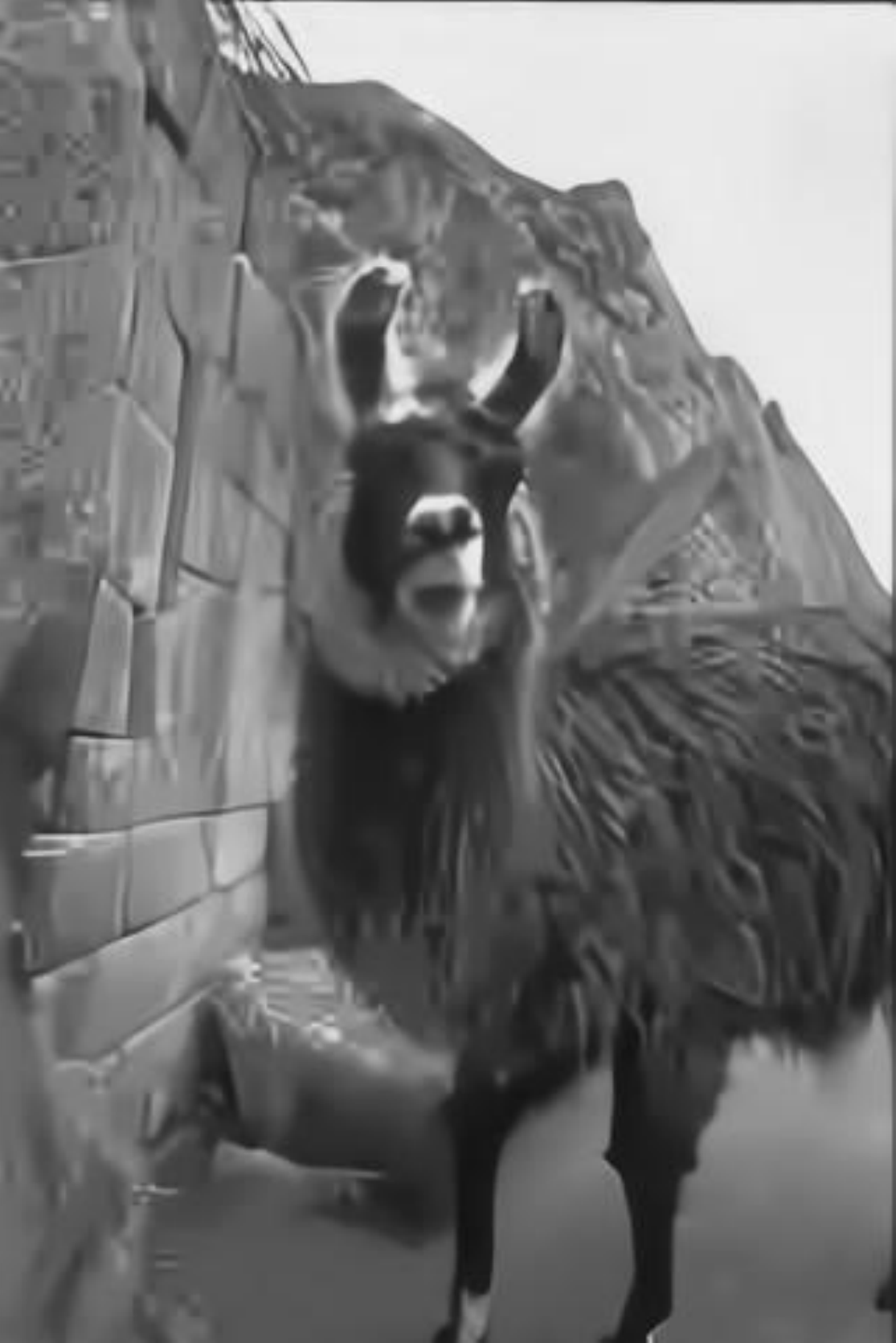}}
\hfil
\subfloat[One-to-Many / $23.59$]{\includegraphics[trim={0 2cm 0 3cm},clip,width=0.19\linewidth]{6046_gan}}
\vspace{-1em}
\subfloat[Ground-truth / PSNR]{\includegraphics[trim={1.5cm 0 1.5cm 0},clip,width=0.19\linewidth]{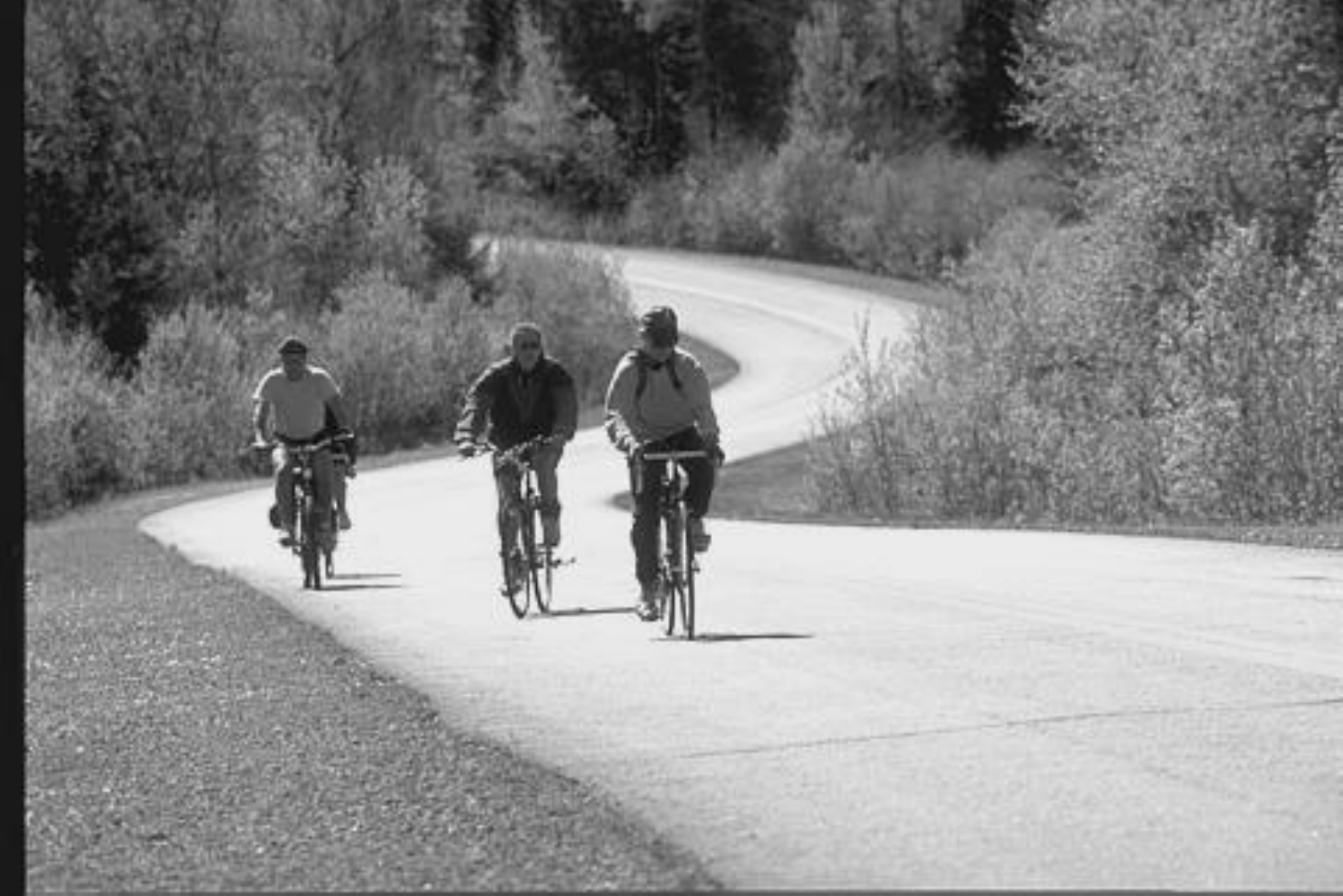}}
\hfil
\subfloat[JPEG / $24.24$]{\includegraphics[trim={1.5cm 0 1.5cm 0},clip,width=0.19\linewidth]{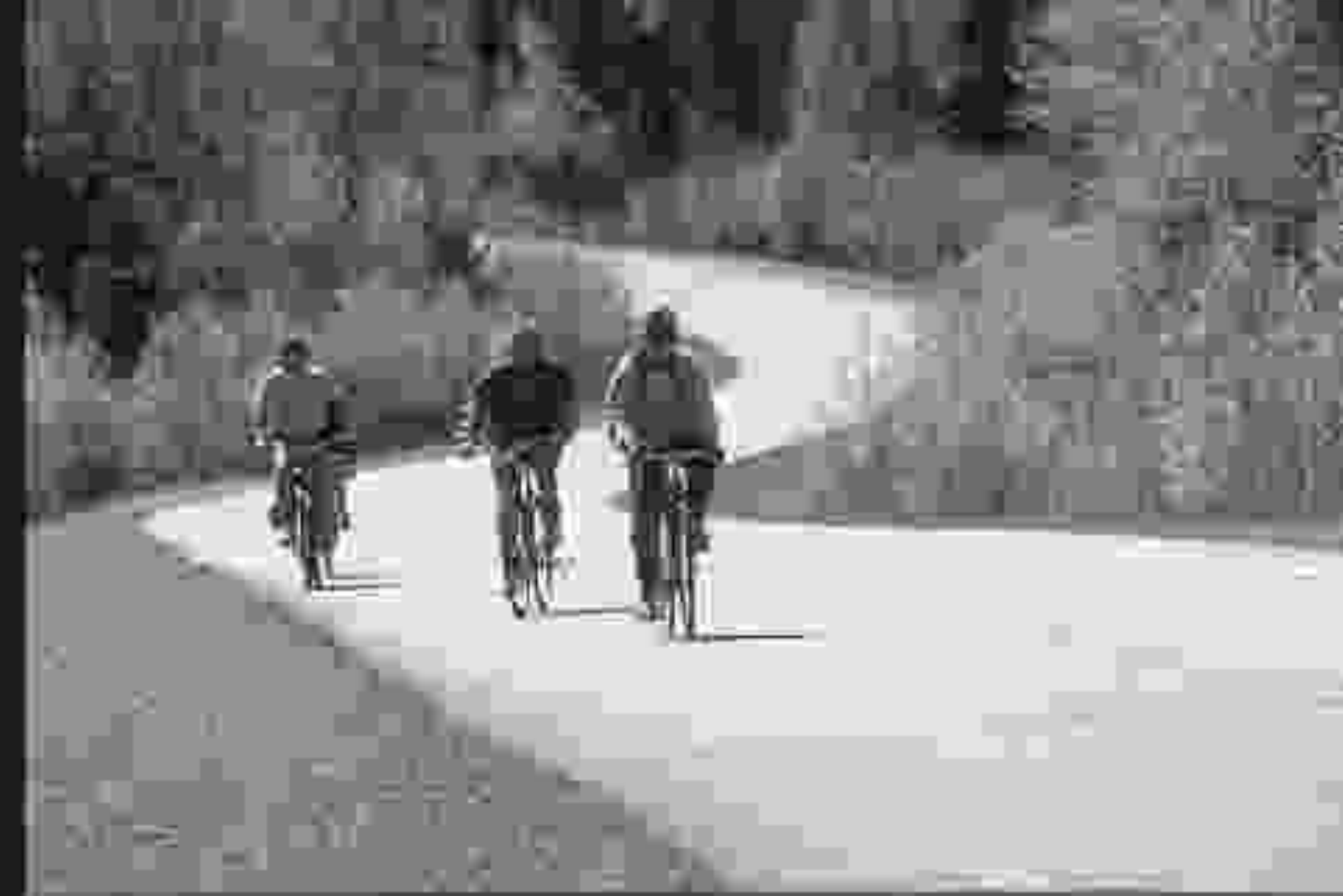}}
\hfil
\subfloat[DDCN / $25.26$]{\includegraphics[trim={1.5cm 0 1.5cm 0},clip,width=0.19\linewidth]{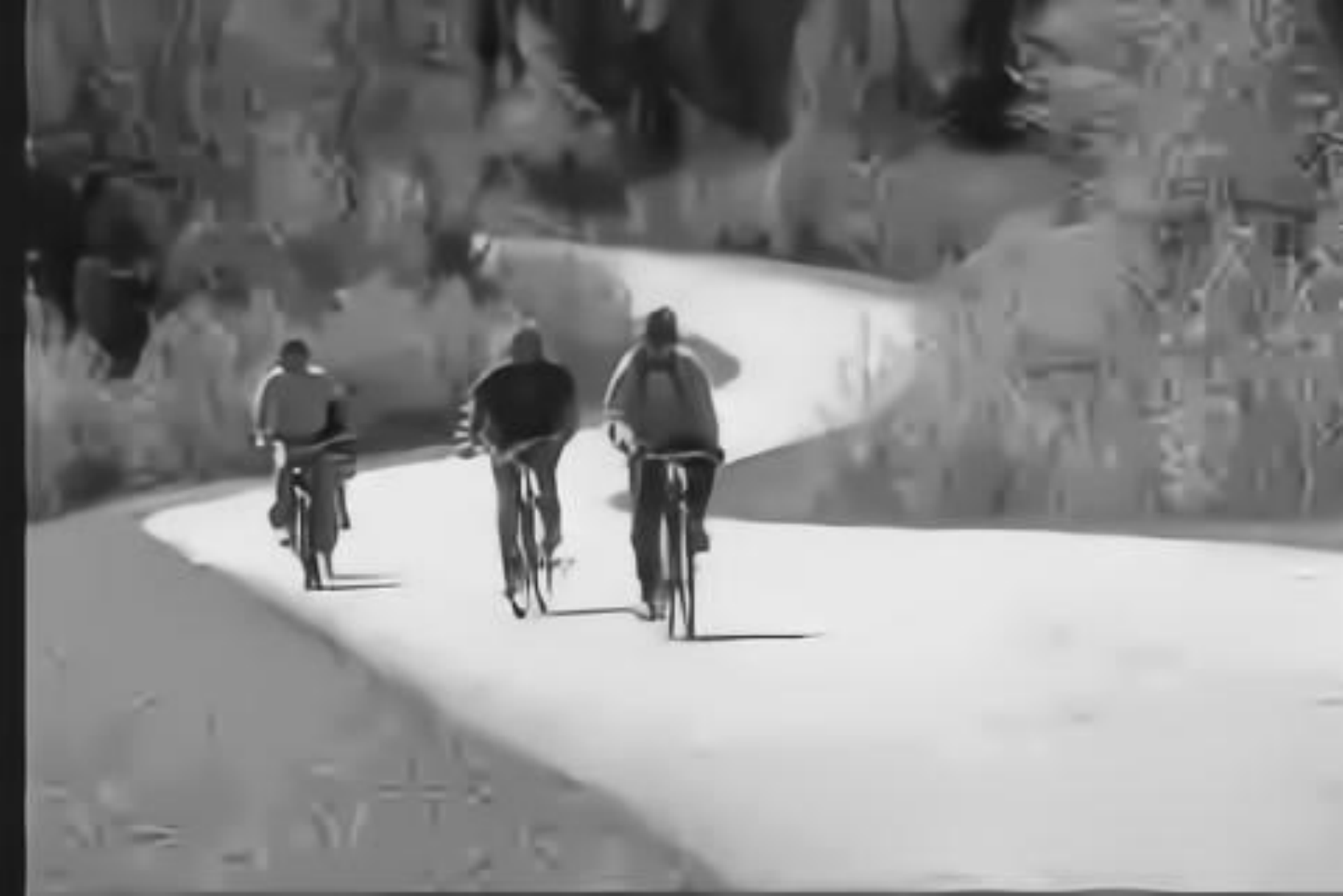}}
\hfil
\subfloat[Baseline / $25.38$]{\includegraphics[trim={1.5cm 0 1.5cm 0},clip,width=0.19\linewidth]{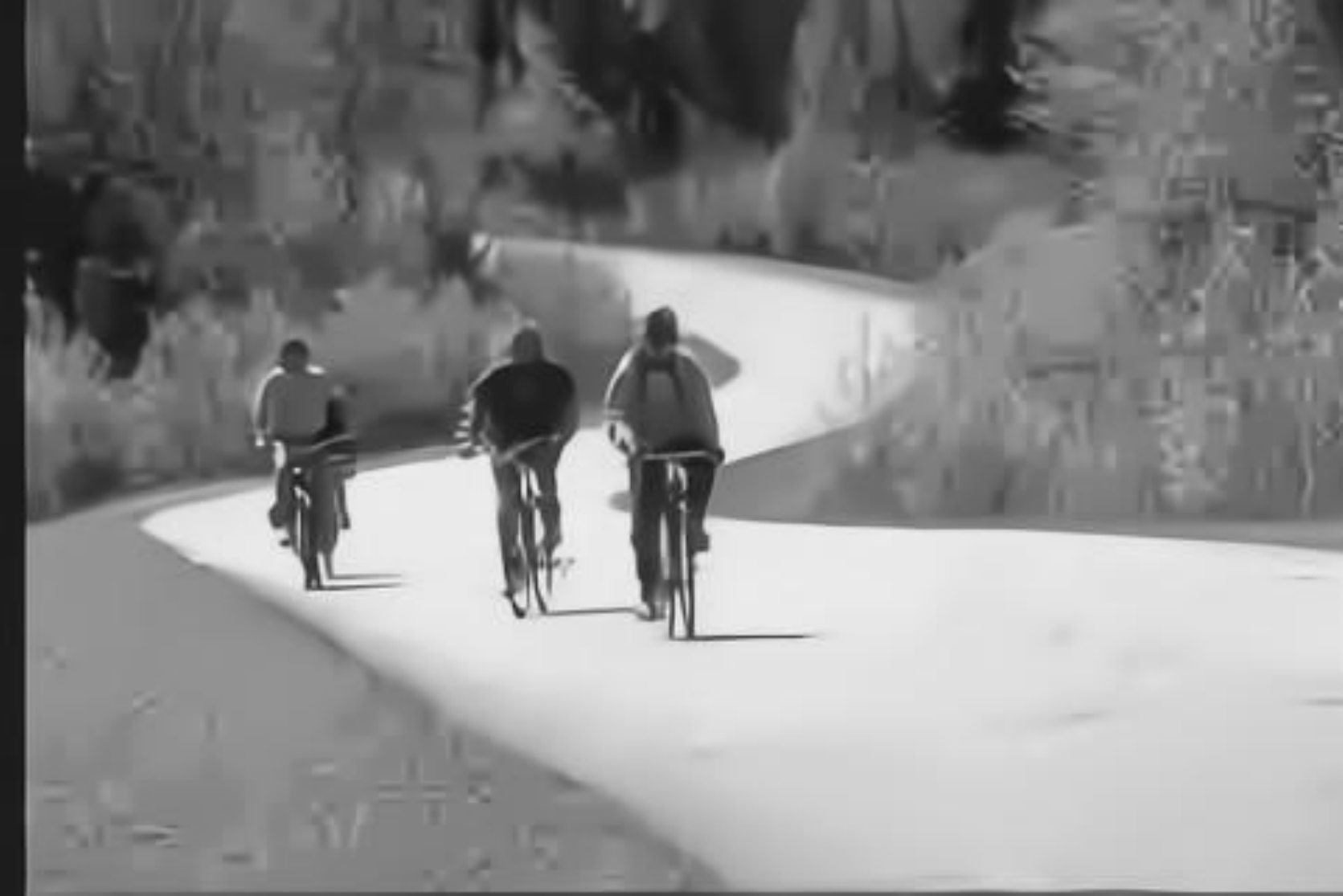}}
\hfil
\subfloat[One-to-Many / $23.57$]{\includegraphics[trim={1.5cm 0 1.5cm 0},clip,width=0.19\linewidth]{344010_gan}}
\vspace{-1em}
\caption{Comparisons under Quality $5$ on BSDS500. Row 1: Image 6046; Row 2: Image 344010. Best view on screen.}
\label{fig:qualitative_bsds}
\vspace{-1.5em}
\end{figure*}

In Cols. 3$\sim$4 of Fig.~\ref{fig:qualitative_bsds}, we present restored images from DDCN and our baseline for qualitative evaluations. In general, we can see that, both approaches tend to produce overly smooth results. The reconstructed images lack fine details and rich textures when compared with the ground-truths. Although our baseline has outperformed existing approaches on PSNR, its visual quality is still far from satisfactory.

\subsection{Favorability Evaluations}
In this section we evaluate the human favorability of recovered images. Unfortunately, currently there is no objective metric to measure the favorability. Hence, we conduct qualitative experiments for evaluations. As has been discussed, the many-to-one quantization step makes recovery highly ill-posed. Ambiguity becomes more extreme as input quality reduces. For low qualities, most high-frequency information of the original image is eliminated. So in the following experiments we focus on Quality $5$ as lower qualities require more effective recovery of details. We train our one-to-many network according to Section \ref{sec:joint_training}. 

The results of our one-to-many network for Image 6045 and Image 344010 are shown in the last column of Fig.~\ref{fig:qualitative_bsds}. By comparing them with the results of existing approaches and our baseline, we can see that our one-to-many network does a pretty good job at recovering edges and details. The textures are much richer in our approach, such as the fur and the rocks in Image 6046, and the bushes in Image 344010, \etc. Note that our one-to-many network does not add details indiscriminately. For example, in Image 6046, our approach largely enriches the fur of the donkey, but the background sky remains clean, suggesting that the one-to-many network is aware of image semantics.

\begin{figure*}[t]
\centering
\captionsetup[subfloat]{labelformat=empty}
\subfloat[Ground-truth / PSNR]{\includegraphics[width=0.19\linewidth]{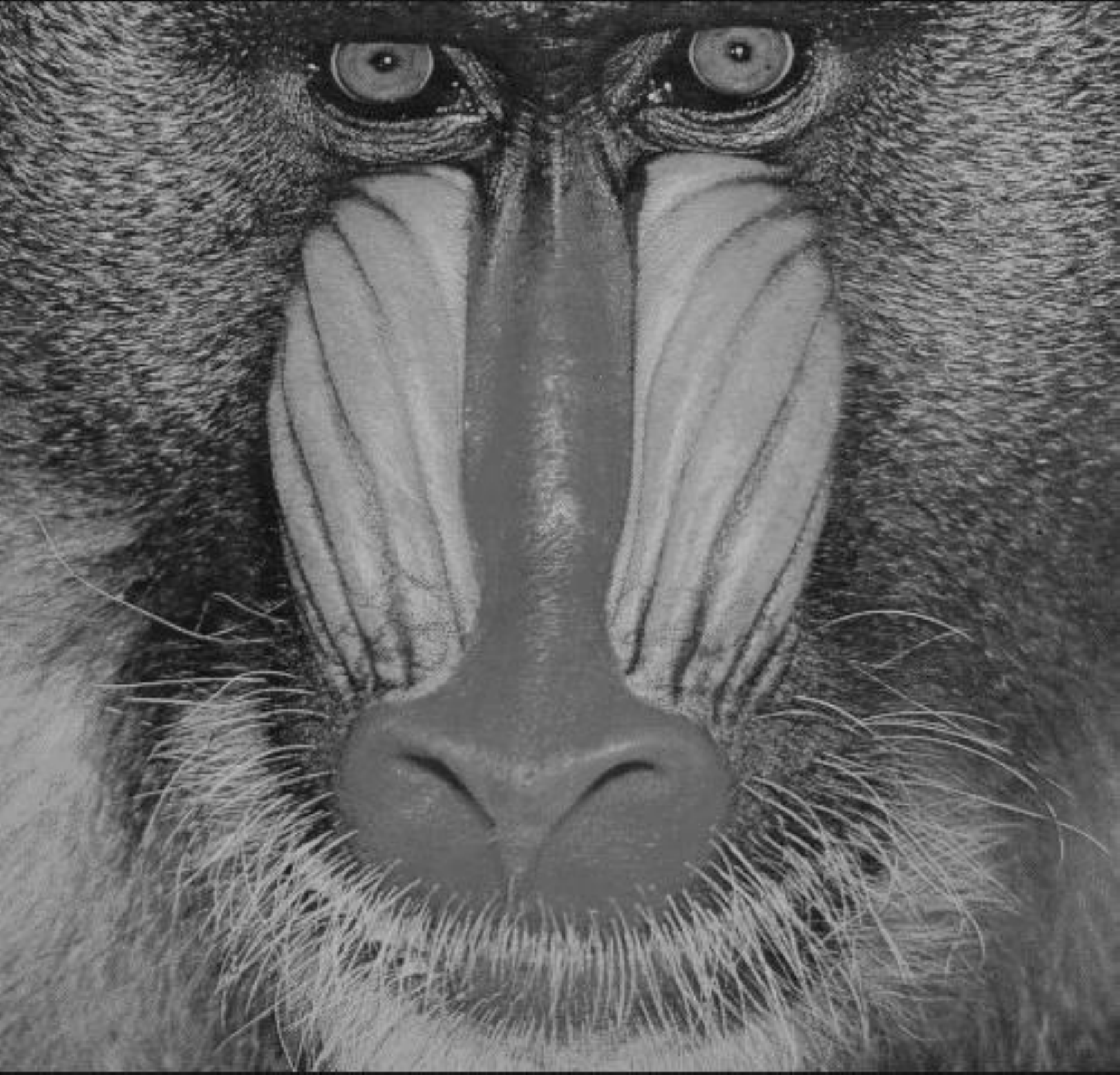}}
\hfil
\subfloat[JPEG / $22.32$]{\includegraphics[width=0.19\linewidth]{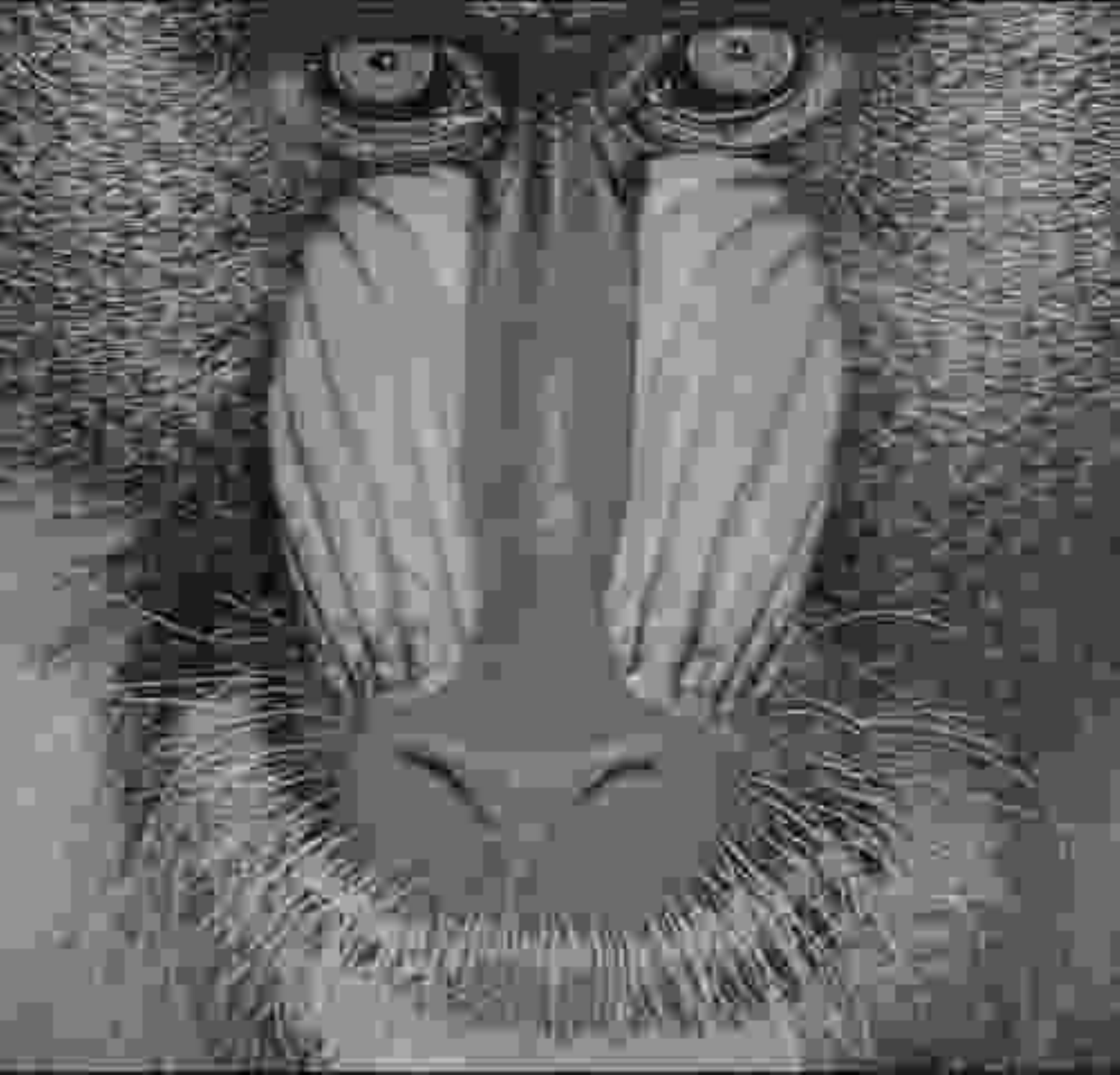}}
\hfil
\subfloat[Baseline / $23.28$]{\includegraphics[width=0.19\linewidth]{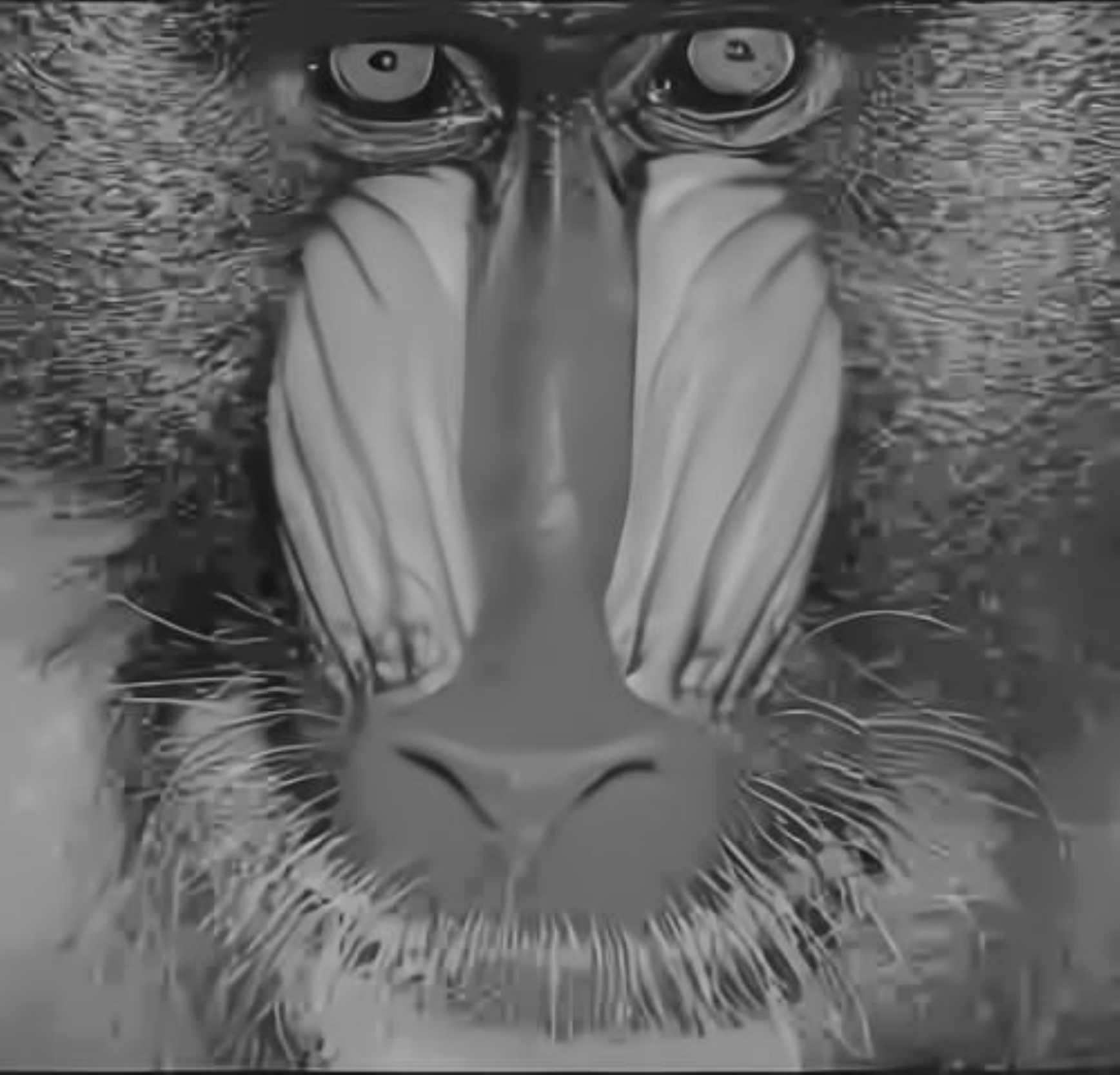}}
\hfil
\subfloat[One-to-Many (1) / $21.50$]{\includegraphics[width=0.19\linewidth]{{{baboon_gan1}}}}
\hfil
\subfloat[One-to-Many (2) / $21.53$]{\includegraphics[width=0.19\linewidth]{{{baboon_gan2}}}}
\vspace{-1em}
\subfloat[Ground-truth / PSNR]{\includegraphics[width=0.19\linewidth]{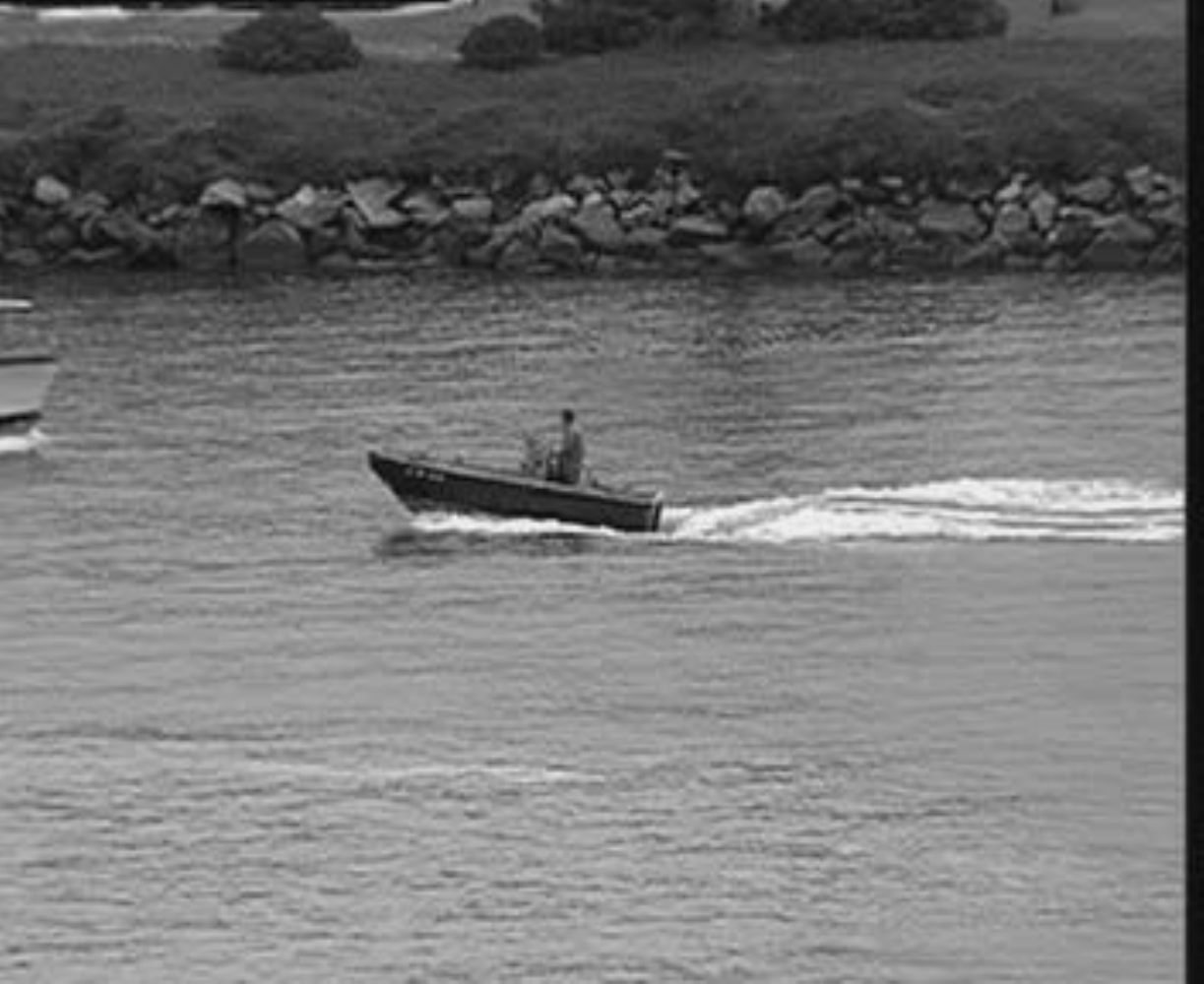}}
\hfil
\subfloat[JPEG / $25.43$]{\includegraphics[width=0.19\linewidth]{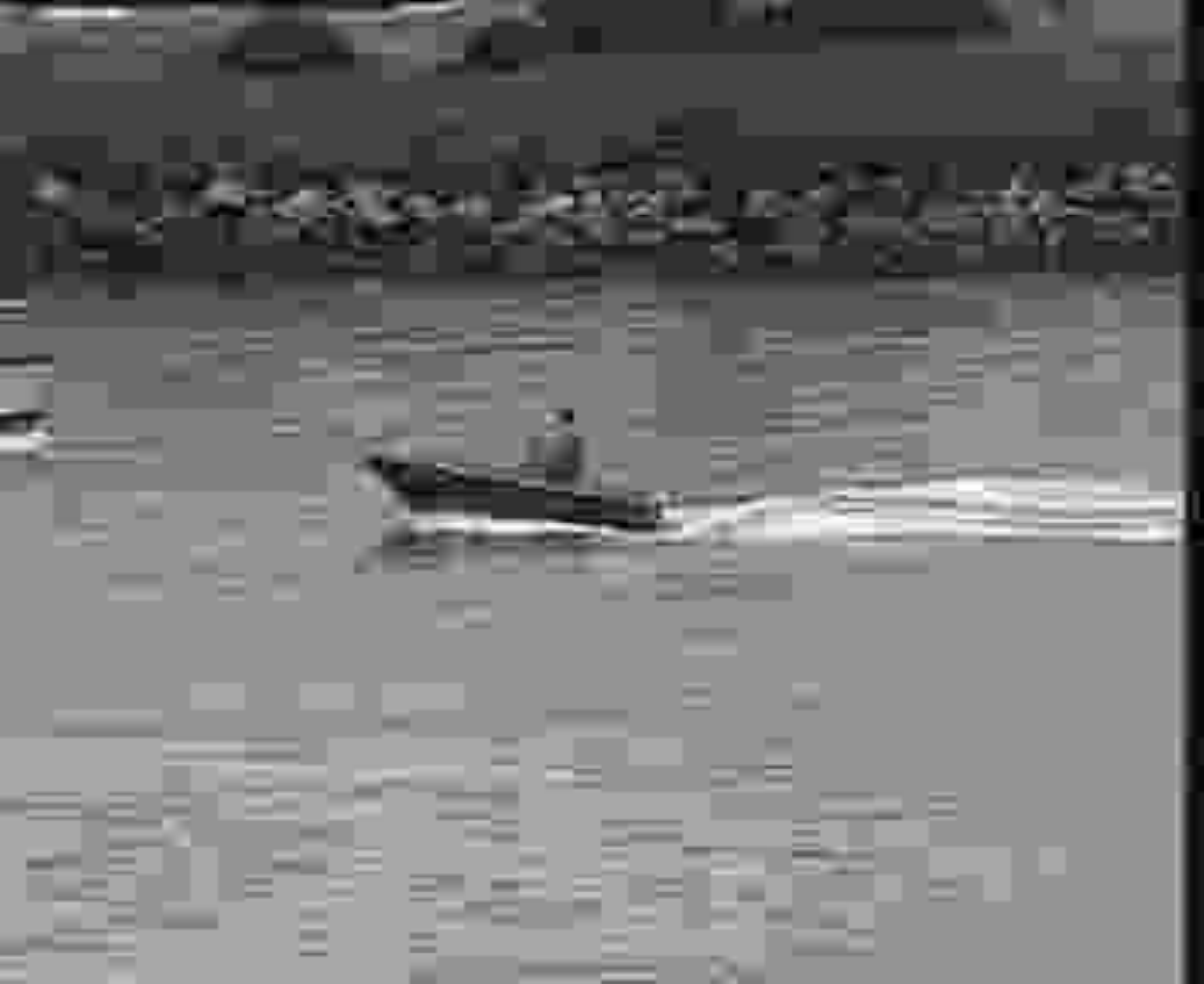}}
\hfil
\subfloat[Baseline / $26.77$]{\includegraphics[width=0.19\linewidth]{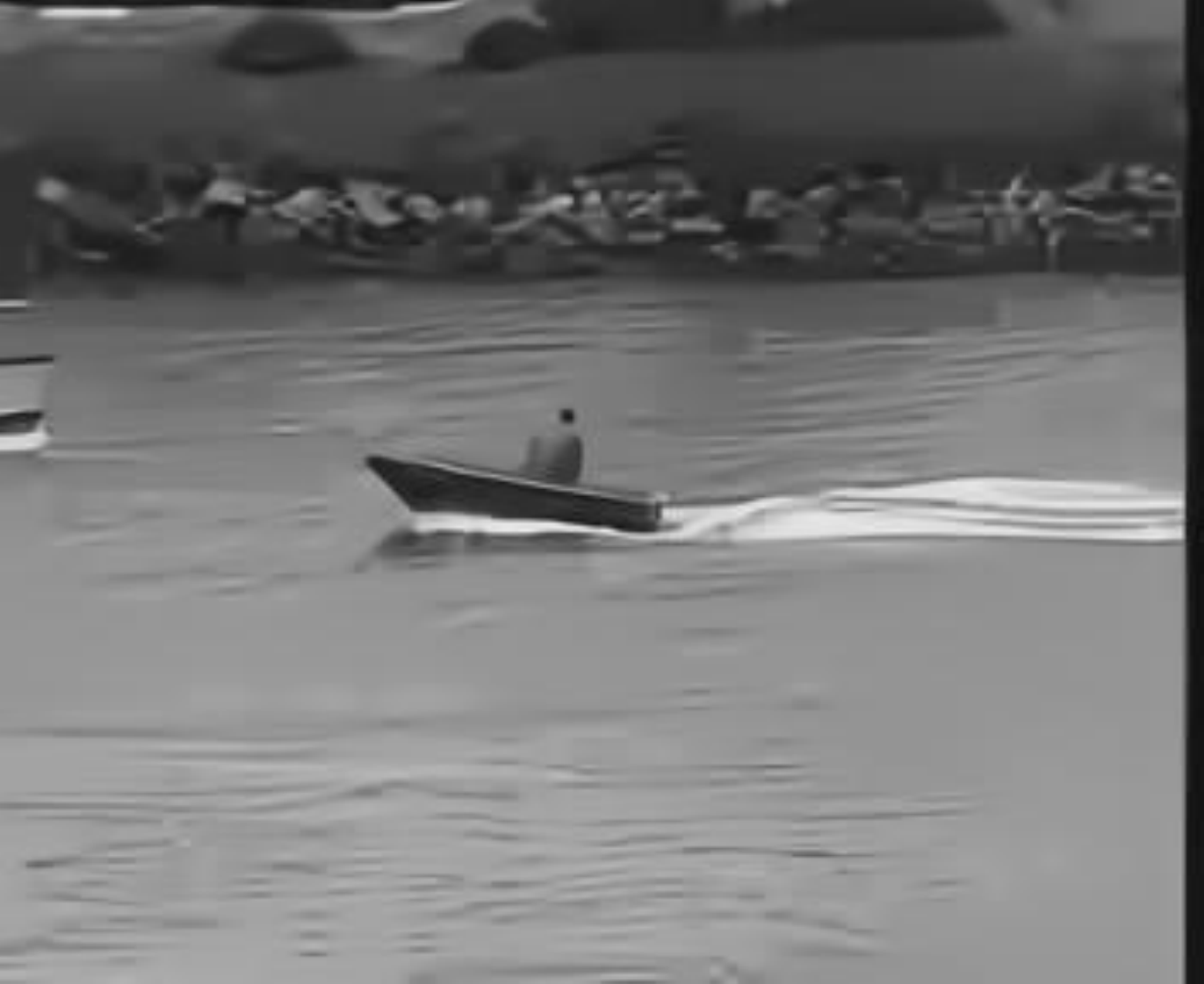}}
\hfil
\subfloat[One-to-Many (1) / $24.99$]{\includegraphics[width=0.19\linewidth]{{{coastguard_gan1}}}}
\hfil
\subfloat[One-to-Many (2) / $25.32$]{\includegraphics[width=0.19\linewidth]{{{coastguard_gan2}}}}
\vspace{-1em}
\caption{Comparison under Quality $5$ on Set14. Row 1: Image ``baboon"; Row 2: Image ``coastguard". Best view on screen.}
\label{fig:qualitative_set14}
\vspace{-1.5em}
\end{figure*}

We also perform evaluations on the Set14 dataset~\cite{set14}. Here, for each compressed image, we sample two different $Z$s to obtain two reconstructed candidates, and show them in the last two columns of Fig.~\ref{fig:qualitative_set14}. As can be seen, these candidates have different details, and all of them look more vivid than the results of our baseline. Interestingly, from the perspective of PSNR, in both Fig.~\ref{fig:qualitative_bsds} and Fig.~\ref{fig:qualitative_set14} our one-to-many network performs even worse than JPEG-compressed input images, whereas it is obvious that our results are more visually pleasing. This demonstrates that PSNR is not sufficient for quality measurement.

\begin{figure*}[t]
\centering
\captionsetup[subfloat]{labelformat=empty}
\subfloat[]{\includegraphics[trim={1.5cm 0 0cm 0},clip,width=0.19\linewidth]{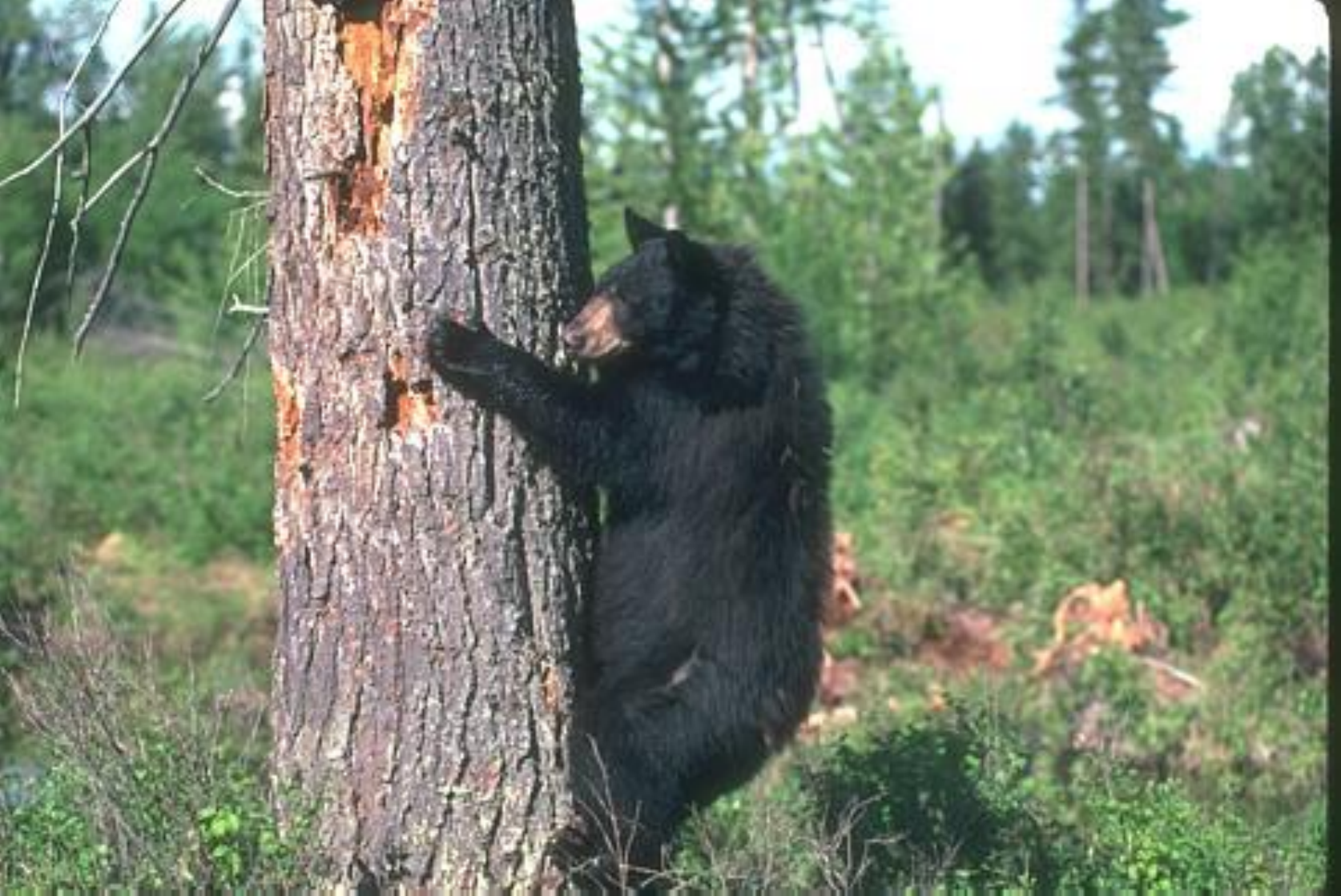}}
\hfil
\subfloat[]{\includegraphics[trim={1.5cm 0 0cm 0},clip,width=0.19\linewidth]{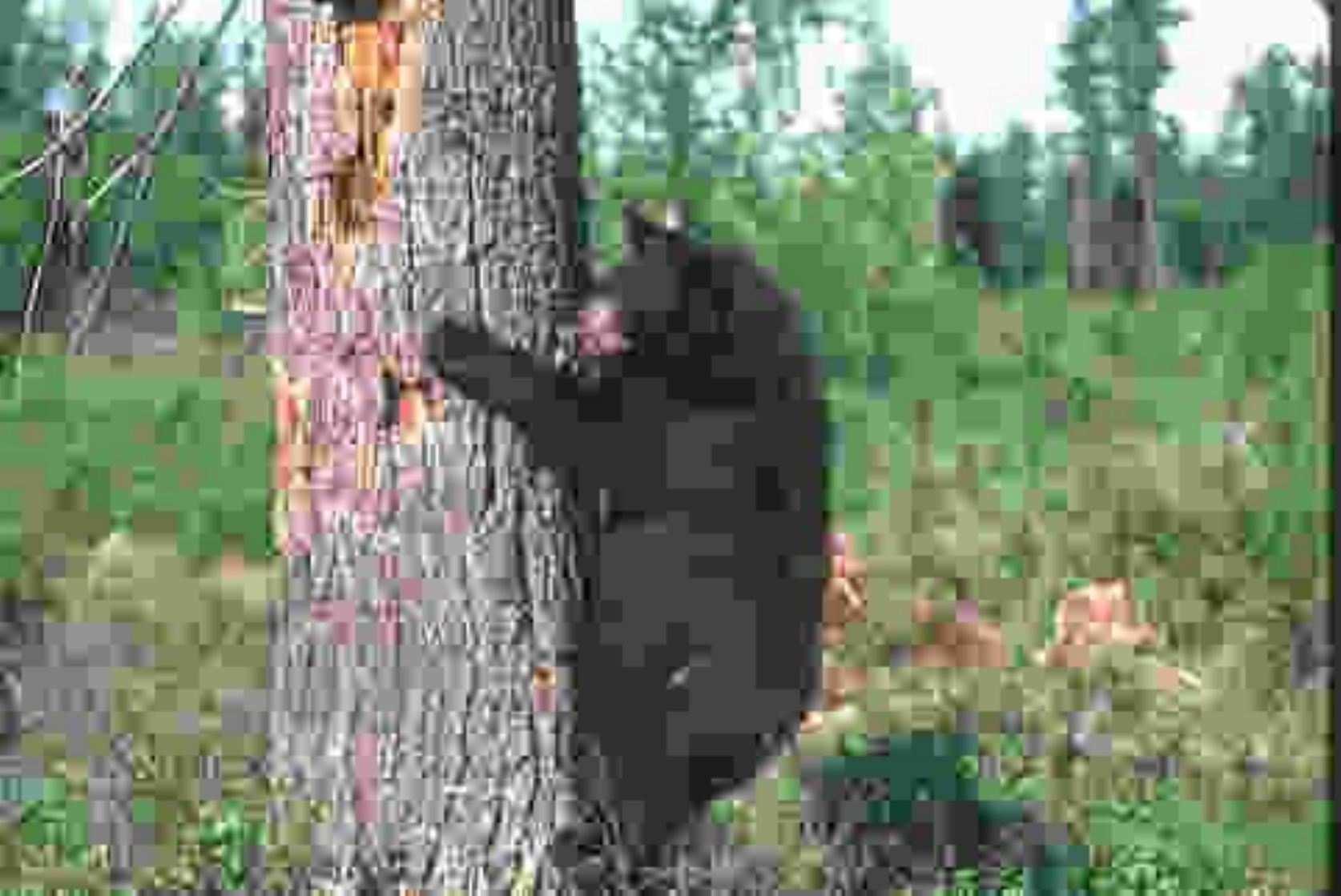}}
\hfil
\subfloat[]{\includegraphics[trim={1.5cm 0 0cm 0},clip,width=0.19\linewidth]{{{100039_ddcn}}}}
\hfil
\subfloat[]{\includegraphics[trim={1.5cm 0 0cm 0},clip,width=0.19\linewidth]{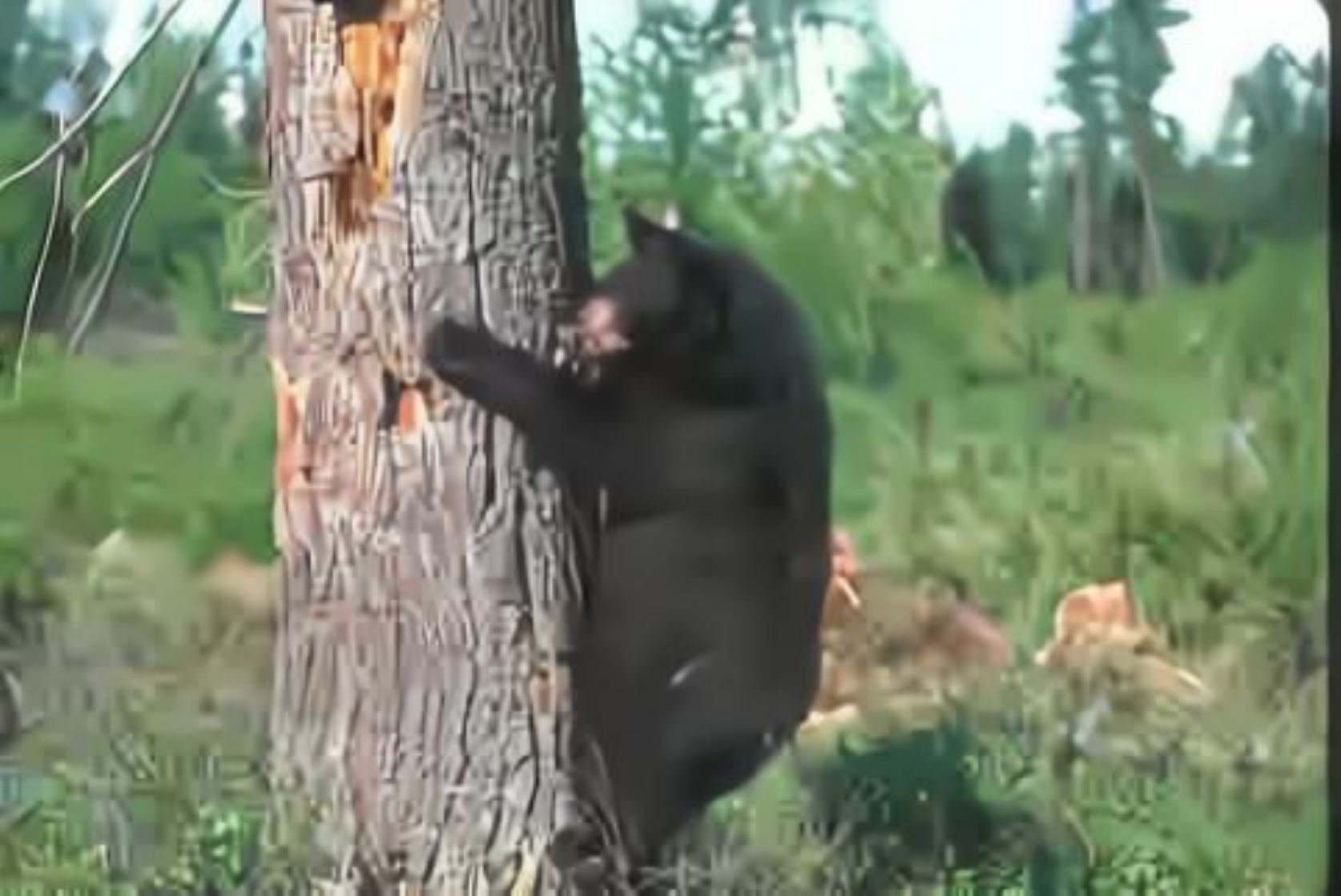}}
\hfil
\subfloat[]{\includegraphics[trim={1.5cm 0 0cm 0},clip,width=0.19\linewidth]{{{100039_gan}}}}
\vspace{-2em}
\subfloat[Ground-truth]{\includegraphics[trim={1.5cm 0 0cm 0},clip,width=0.19\linewidth]{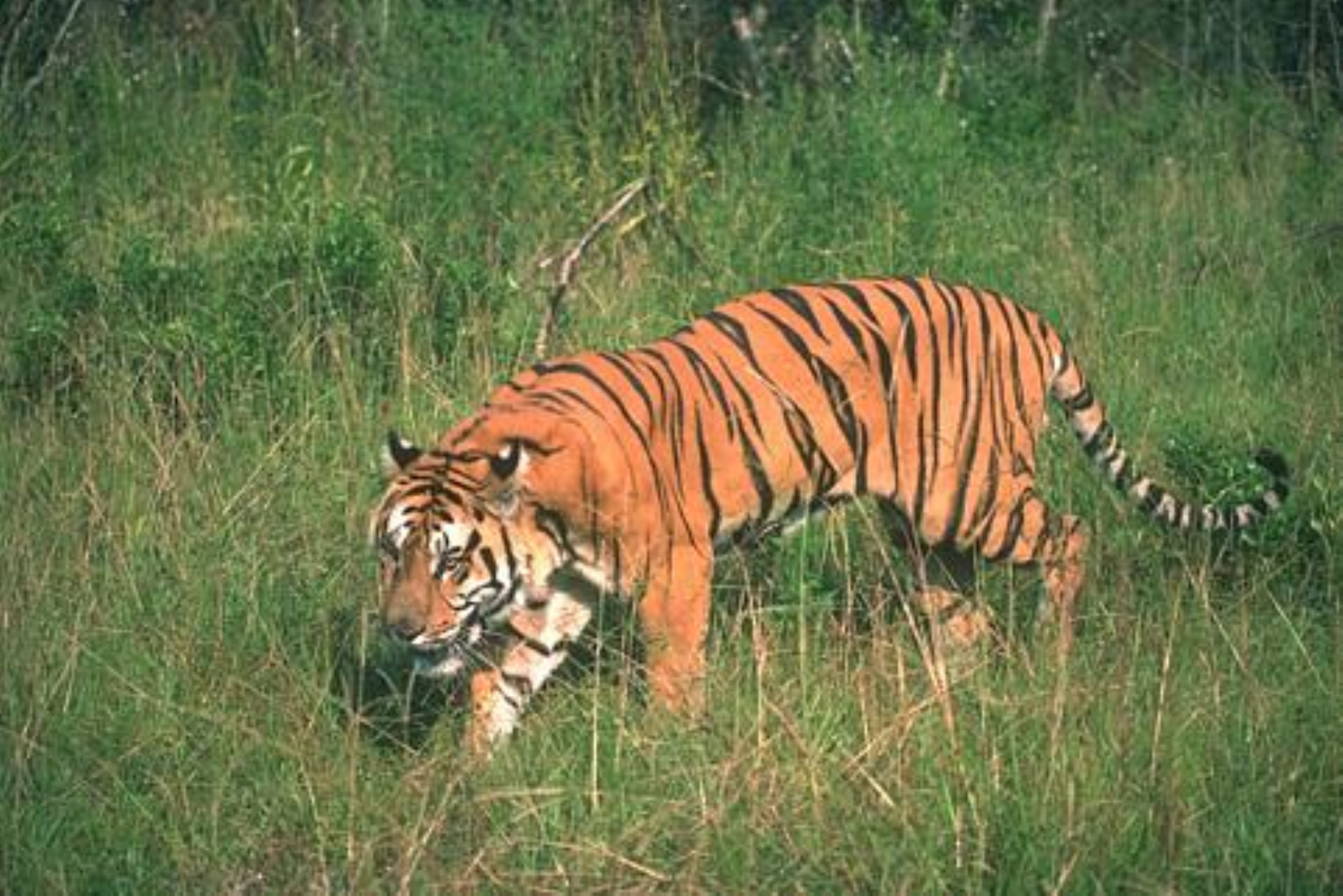}}
\hfil
\subfloat[JPEG]{\includegraphics[trim={1.5cm 0 0cm 0},clip,width=0.19\linewidth]{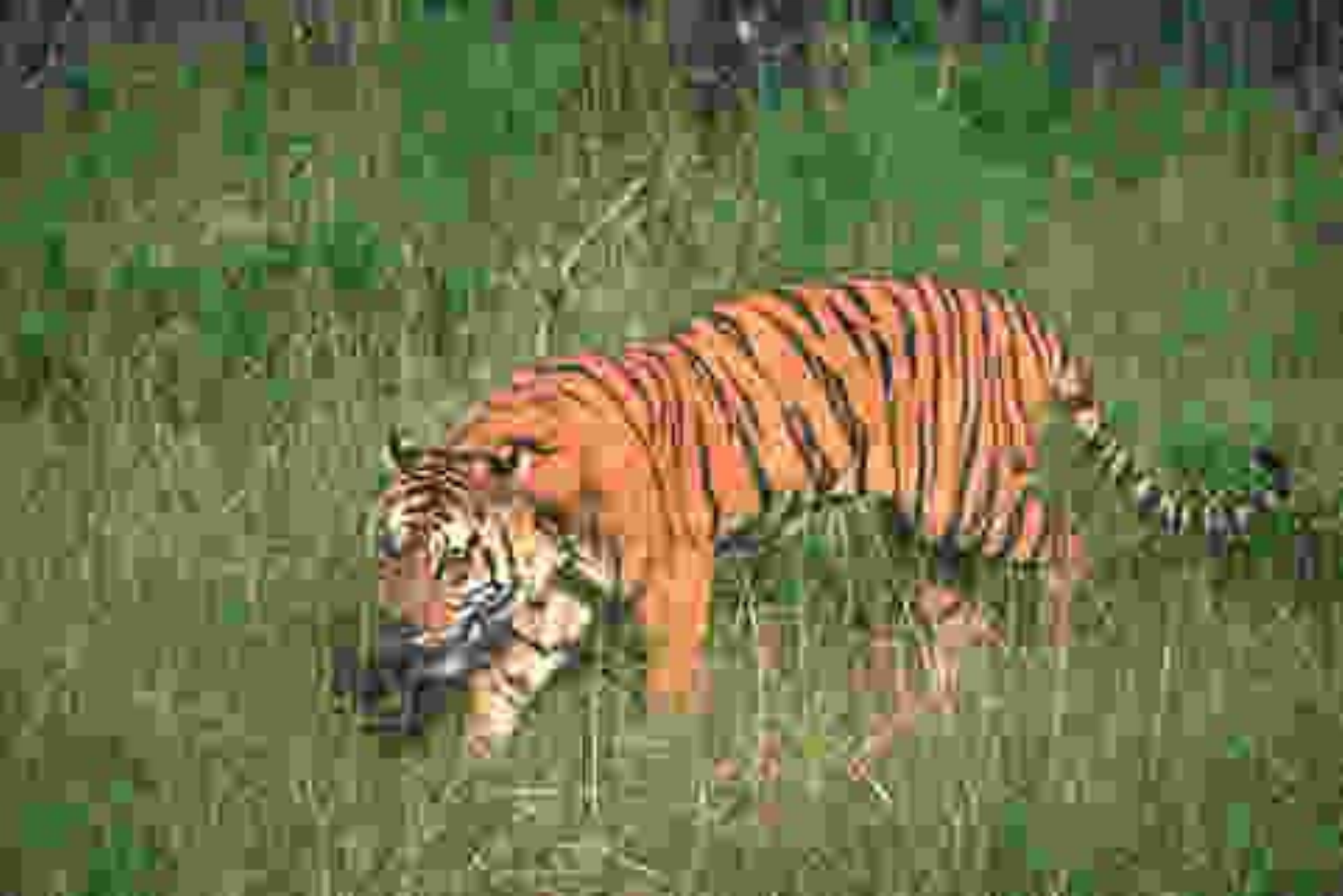}}
\hfil
\subfloat[DDCN]{\includegraphics[trim={1.5cm 0 0cm 0},clip,width=0.19\linewidth]{{{108004_ddcn}}}}
\hfil
\subfloat[Baseline]{\includegraphics[trim={1.5cm 0 0cm 0},clip,width=0.19\linewidth]{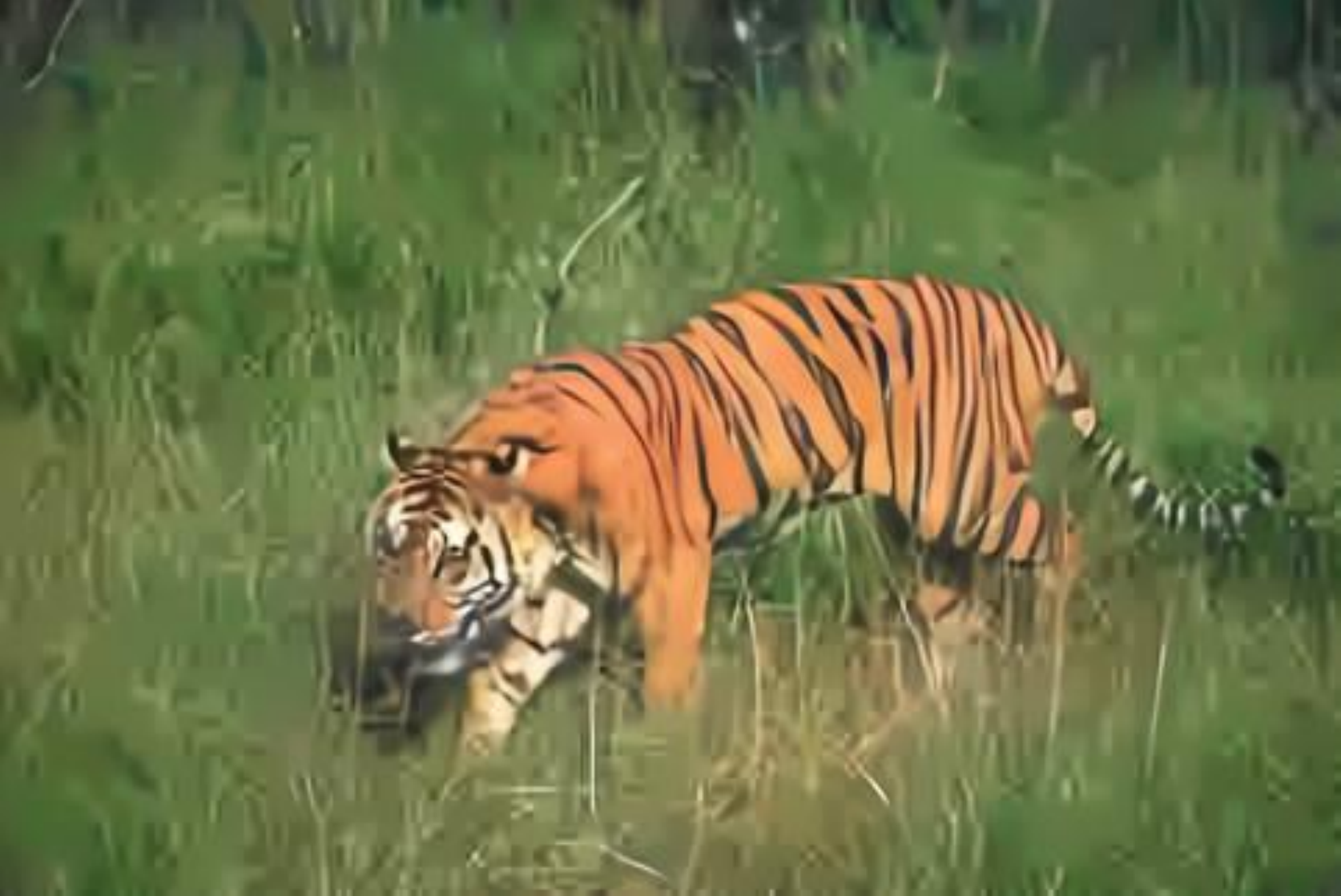}}
\hfil
\subfloat[One-to-Many]{\includegraphics[trim={1.5cm 0 0cm 0},clip,width=0.19\linewidth]{{{108004_gan}}}}
\vspace{-1em}
\caption{Comparison under Quality $5$ on BSDS500. Row 1: Image 100039; Row 2: Image 108004. Best view on screen.}
\label{fig:color}
\vspace{-1.5em}
\end{figure*}

\subsection{Extension to Color Images}
Our approach is not limited to gray images. We re-train our one-to-many network on RGB images and show results in Fig.~\ref{fig:color}. As can be observed, our approach has produced much finer details in comparison to DDCN or our baseline.

\subsection{Further Analysis}
\begin{figure}[t]
\vspace{-0.5em}
\centering
\subfloat[]{
\includegraphics[width=0.3\columnwidth]{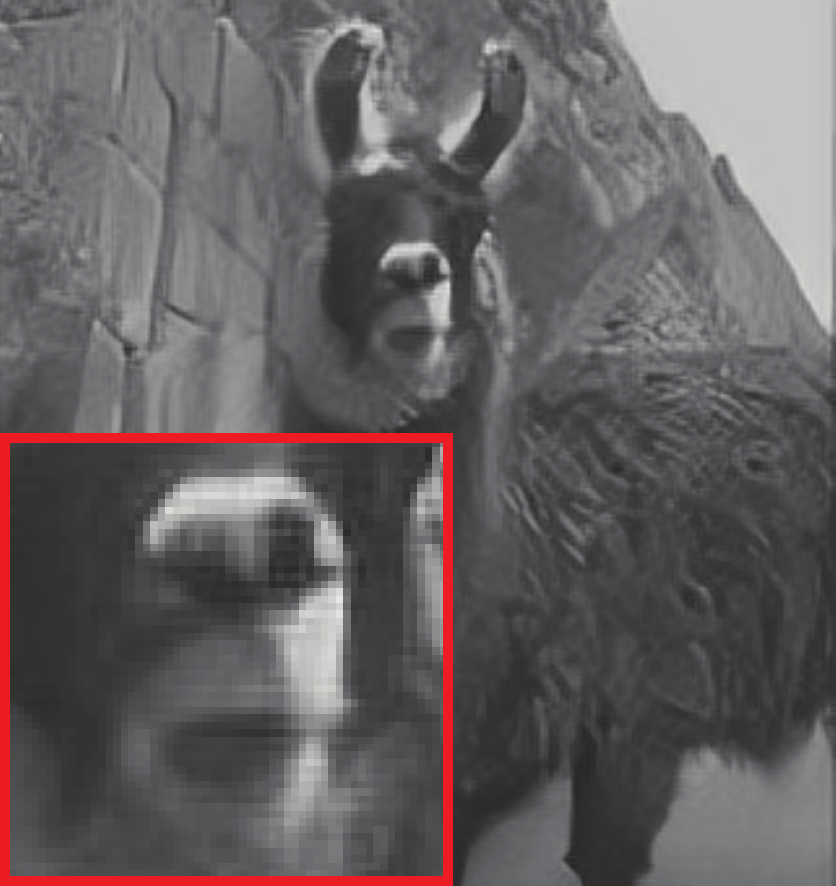}
\label{fig:percept}}
\hfil
\subfloat[]{
\includegraphics[width=0.3\columnwidth]{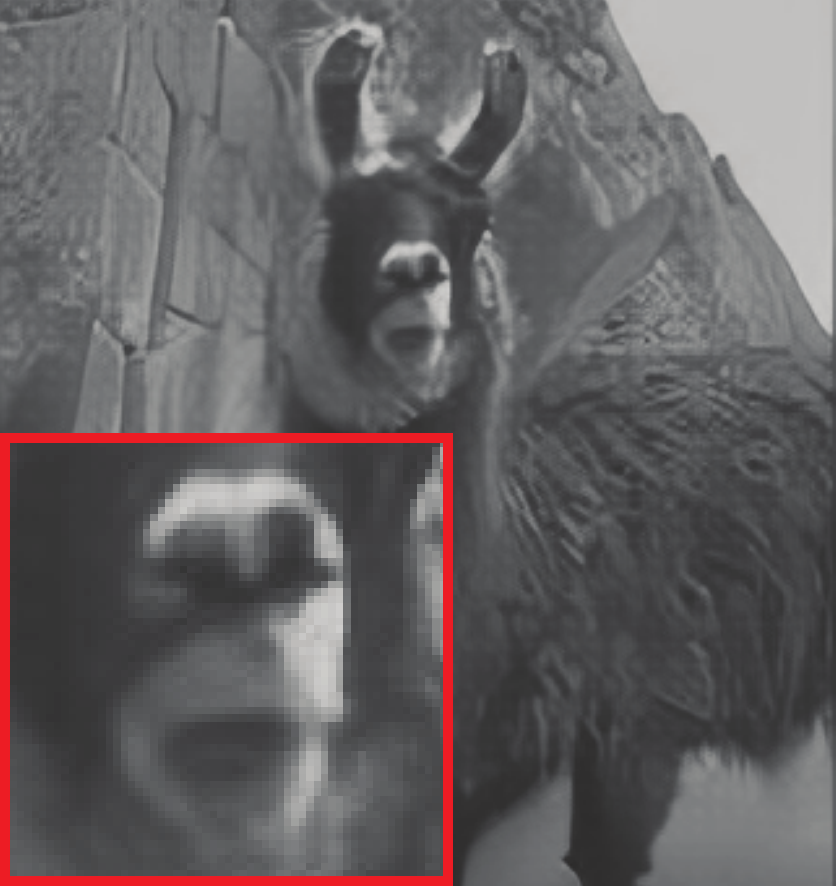}
\label{fig:percept_shift}}
\hfil
\subfloat[]{
\includegraphics[width=0.3\columnwidth]{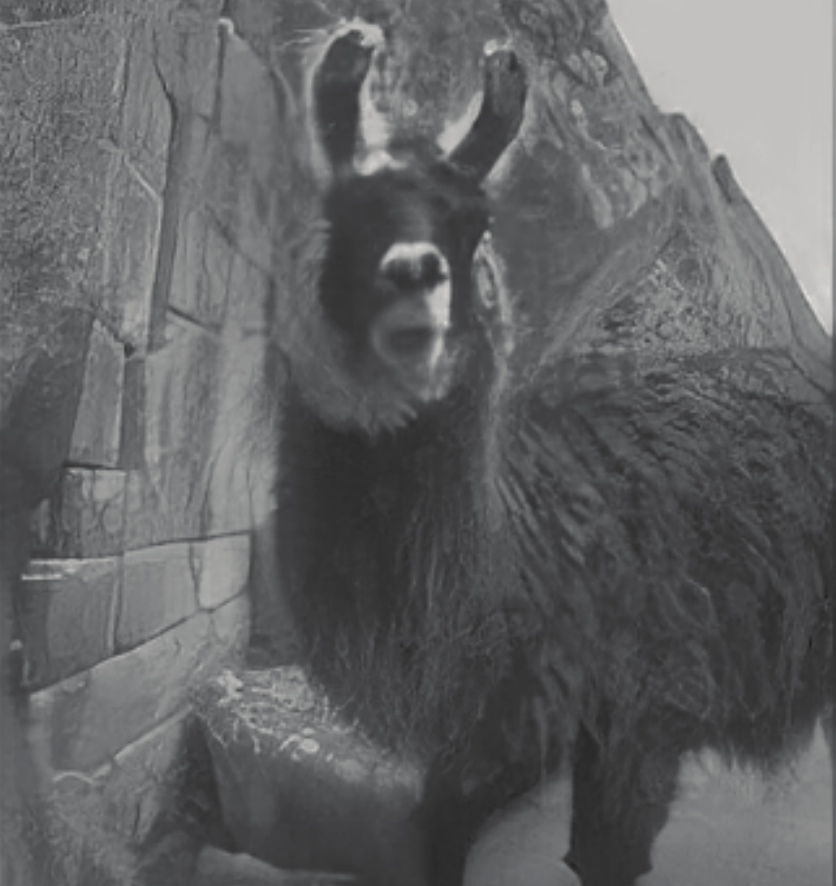}
\label{fig:percept_natural}}
\vspace{-1em}
\caption{(a) Train using $L_{percept}$ without the shift-and-average strategy; (b) Train using $L_{percept}$ with the shift-and-average strategy; (c) Train using $L_{percept}$ and $L_{natural}$.}
\label{fig:further_analysis}
\vspace{-1.5em}
\end{figure}

\paragraph{Up-sample}
We conduct another experiment to examine the proposed shift-and-average strategy. By comparing Fig.~\ref{fig:percept} and Fig.~\ref{fig:percept_shift}, we can observe that, without the proposed strategy, grid-like artifacts are visible over the whole image, suggesting that they are results of the composition of a traditional deconvolution operation and a highly non-convex loss function. The proposed strategy is able to suppress such artifacts without hurting the perceptual quality.

\paragraph{Loss}
Fig.~\ref{fig:further_analysis} also exhibits the influences of each loss. Fig.~\ref{fig:percept_shift} shows that using $L_{percept}$ for measurement is able to recover primary semantic information. After adding $L_{natural}$, fine details are supplemented, as indicated in Fig.~\ref{fig:percept_natural}. However, the contrast of the resulting image is significantly different from the input or ground-truth (both can be found in Fig.~\ref{fig:qualitative_bsds}). Once we put in $L_{jpeg}$, the contrast is adjusted (see Row 1, Column 5 of Fig.~\ref{fig:qualitative_bsds}). More interestingly, it seems that additional details appear as well. One explanation is, high-contrast natural images usually have more complex textures than low-contrast images, so $L_{natural}$ will encourage the network to synthesize more details after the contrast is enhanced by $L_{jpeg}$. These experiments demonstrate the importance of all proposed losses.

\section{Conclusion}
In this paper, we systematically studied how to effectively recover artifact-free images from JPEG-compressed images. As a natural inversion of the many-to-one JPEG compression, we propose a one-to-many network. The proposed model, when optimized with a perceptual loss, a naturalness loss, and a JPEG loss, could reconstruct multiple artifact-free candidates that are more favored by humans, and thus drastically improved the recovery quality.

One limitation of our approach is the scalability to various JPEG qualities. Currently we need to train an individual model for each quality. Besides, how to objectively evaluate the favorablity of output images remains to be a problem. We hope to address these issues in the future.

\section*{Acknowledgment}
This work is supported by NSF of China under Grant U1611461, 61672548, and the Guangzhou Science and Technology Program, China, under Grant 201510010165.

{\small
\bibliographystyle{ieee}
\bibliography{egbib}
}

\end{document}